\documentclass[pdflatex,sn-mathphys-num]{sn-jnl}% Math and Physical Sciences Numbered Reference Style
\usepackage{graphicx}%
\usepackage{multirow}%
\usepackage{amsmath,amssymb,amsfonts}%
\usepackage{amsthm}%
\usepackage{mathrsfs}%
\usepackage[title]{appendix}%
\usepackage{xcolor}%
\usepackage{textcomp}%
\usepackage{manyfoot}%
\usepackage{booktabs}%
\usepackage{algorithm}%
\usepackage{algorithmicx}%
\usepackage{algpseudocode}%
\usepackage{listings}%
\usepackage{bibunits}%
\usepackage[hypcap=true]{caption}

\graphicspath{{imgs/}{schematic-plots/}}
\theoremstyle{thmstyleone}%
\theoremstyle{thmstyletwo}%

\theoremstyle{thmstylethree}%

\algrenewcommand\algorithmicrequire{\textbf{Input:}}
\algrenewcommand\algorithmicensure{\textbf{Output:}}
\algnewcommand{\LeftComment}[1]{\Statex \(\triangleright\) #1}

\newcommand{\reals}{\mathbb{R}}

\begin{document}

\title[Article Title]{Fast and principled equation discovery from chaos to climate}


\author[1]{\fnm{Yuzheng} \sur{Zhang}}

\author[3,1]{\fnm{Weizhen} \sur{Li}}%\email{weizhen.li@dur.ac.uk}

\author*[1,2]{\fnm{Rui} \sur{Carvalho}}\email{rui.carvalho@dur.ac.uk}

\affil[1]{\orgdiv{Department of Engineering}, \orgname{Durham University}, \orgaddress{\street{Lower Mountjoy, South Road}, \city{Durham}, \postcode{DH1~3LE}, \country{United Kingdom}}}

\affil[2]{\orgdiv{Institute for Data Science}, \orgname{Durham University}, \orgaddress{\street{South Road}, \city{Durham}, \postcode{DH1~3LE}, \country{United Kingdom}}}

\affil[3]{\orgdiv{College of Control Science and Engineering}, \orgname{Zhejiang University}, \orgaddress{\street{38 Zheda Rd.(Yuquan Campus)}, \city{Hangzhou}, \postcode{310027}, \country{P.R. China}}}


\abstract{
    Our ability to predict, control, and ultimately
    understand complex systems rests on discovering the equations that
    govern their dynamics. Identifying these equations directly from
    noisy, limited observations has therefore become a central
    challenge in data-driven science, yet existing library-based
    sparse regression methods force a compromise between automation,
    statistical rigor, and computational efficiency. Here we develop Bayesian-ARGOS, a hybrid framework that reconciles these demands by combining rapid frequentist screening with focused Bayesian inference, enabling automated equation discovery with principled uncertainty quantification at a fraction of the computational cost of existing methods. Tested on seven chaotic systems under varying data scarcity and noise levels, Bayesian-ARGOS outperforms two state-of-the-art methods in most scenarios. It surpasses SINDy in data efficiency for all systems and noise tolerance for six out of the seven, with a two-order-of-magnitude reduction in computational cost compared to bootstrap-based ARGOS. The probabilistic formulation additionally enables a suite of standard statistical diagnostics, including influence analysis and multicollinearity detection that expose failure modes otherwise opaque. When integrated with representation learning (SINDy-SHRED) for high dimensional sea surface temperature reconstruction, Bayesian-ARGOS increases the yield of valid latent equations with significantly improved long horizon stability. Bayesian-ARGOS thus provides a principled, automated, and computationally efficient route from scarce and noisy observations to interpretable governing equations, offering a practical framework for equation discovery across scales, from benchmark chaotic systems to the latent dynamics underlying global climate patterns.
}

% In systematic evaluations probing distinct limits of data scarcity and noise intensity across seven chaotic systems, Bayesian-ARGOS outperforms two state-of-the-art methods in the majority of cases. It surpasses SINDy in data efficiency for all systems and noise tolerance for six (surpassing ARGOS in five and four, respectively), with a two-order-of-magnitude reduction in computational cost compared to bootstrap-based ARGOS. 




\maketitle

\section{Introduction}
\label{sec:introduction}

Since Newton's time, scientific discovery has predominantly followed two distinct paradigms: the Keplerian (data-driven) approach, which extracts patterns from empirical data as exemplified by planetary laws, and the Newtonian (first-principles) approach, which derives fundamental governing principles through theoretical reasoning, ranging from classical motion to Schrödinger's quantum mechanics~\cite{eDawningNewEra2021}. The recent proliferation of sensor data and computational resources has vastly amplified data-driven methods; however, while these techniques excel at recognizing complex patterns, they often lack mechanistic interpretability. First-principles approaches, by contrast, provide interpretable governing equations, yet their derivation remains a labor-intensive process that demands deep domain expertise and resists automation. Data-driven model discovery synthesizes these paradigms: it leverages the pattern-recognition power of modern algorithms while preserving the foundational insights of theory by identifying parsimonious governing equations directly from empirical data.

Over the past decade, a powerful paradigm has emerged for this task: approximate the unknown dynamics of an ODE or PDE as a linear combination of candidate basis functions and use sparsity-promoting regression to identify the significant terms from data~\cite{bruntonDiscoveringGoverningEquations2016, rudyDatadrivenDiscoveryPartial2017}. This library-based sparse regression approach has yielded parsimonious, interpretable models across domains from fluid mechanics to neuroscience~\cite{gaoDatadrivenInferenceComplex2023, bruntonPromisingDirectionsMachine2024}. Its seamless integration with neural networks further broadens its applicability: by combining the representational power of deep learning with the interpretability of sparse equations, it enables both reduced-order modeling of high-dimensional observations~\cite{championDatadrivenDiscoveryCoordinates2019, rautArousalUniversalEmbedding2025} and surrogate modeling of complex network dynamics~\cite{gaoAutonomousInferenceComplex2022}, underscoring the central role of library-based sparse regression in modern equation discovery pipelines.

Despite this progress, library-based methods still face a fundamental tension among three goals: automation (minimal manual tuning), statistical rigor (principled model selection with uncertainty quantification), and computational efficiency. Existing approaches have advanced individual aspects of this trilemma, but none reconcile all three simultaneously. Within the frequentist paradigm, successive refinements make this trade-off clear. Deterministic thresholding methods are computationally efficient, yet they lack principled selection criteria and depend heavily on manually chosen hyperparameters~\cite{bruntonDiscoveringGoverningEquations2016, zhengUnifiedFrameworkSparse2019}. Augmenting sparse regression with information criteria provides a more principled basis for model selection and reduces ad hoc threshold choices~\cite{manganModelSelectionDynamical2017}, but leaves coefficient uncertainty unquantified. Ensemble and bootstrap extensions partially address this gap by supplying inclusion frequencies or confidence intervals, at the cost of repeated refitting and increased computation~\cite{faselEnsembleSINDyRobustSparse2022}. Recent automated frequentist pipelines combine adaptive lasso, information criteria, and bootstrap confidence intervals, substantially reducing manual tuning while improving statistical rigor, but their multiple resampling stages still incur substantial computational cost~\cite{eganAutomaticallyDiscoveringOrdinary2024}. Bayesian approaches offer a complementary route. In a Bayesian formulation, marginal-likelihood optimization supports principled model comparison through an Occam factor that penalizes unnecessary complexity~\cite{mackayBayesianInterpolation1992}, while posterior distributions provide a natural basis for quantifying parameter uncertainty and can improve robustness under noise and limited data~\cite{panSparseBayesianApproach2016, zhangRobustDatadrivenDiscovery2018, champneysBINDyBayesianIdentification2025b}. However, fully Bayesian posterior inference via MCMC scales poorly to the large candidate libraries typical of system identification~\cite{martina-perezBayesianUncertaintyQuantification2021}, whereas faster evidence-maximization methods often rely on Gaussian noise models and hierarchical priors whose approximations can become unreliable in strongly correlated libraries or under more complex posterior structure~\cite{tippingFastMarginalLikelihood2003}. Existing methods therefore still do not satisfactorily reconcile all three goals.

Here we develop Bayesian-ARGOS, a hybrid framework that reconciles this tension by strategically decomposing the discovery process into two complementary stages: a frequentist screening stage that aggressively reduces the candidate library, followed by a Bayesian stage that performs posterior inference over the reduced model space. The screening stage uses a dual-pass sequential regression procedure combining cross-validated adaptive lasso, ordinary least squares (OLS), and information selection criteria to achieve rapid dimensionality reduction. This division of labor leverages complementary strengths: the adaptive lasso enables efficient library reduction despite shrinkage bias, while Bayesian inference provides principled uncertainty quantification but scales poorly with large libraries. By applying each method where it performs optimally, Bayesian-ARGOS achieves three key advantages: reduced manual tuning, principled model selection via posterior distributions, and confinement of expensive MCMC sampling to a tractable, pre-screened candidate set.

We validate Bayesian-ARGOS through comprehensive benchmarking against ARGOS and SINDy across seven chaotic systems recommended for evaluating system identification algorithms~\cite{kaptanogluBenchmarkingSparseSystem2023}, systematically varying observation counts and noise levels. In data efficiency, Bayesian-ARGOS achieves consistent identification with fewer observations than SINDy across all tested systems while outperforming ARGOS in five of seven cases. For noise robustness, it demonstrates superior tolerance compared to ARGOS in four systems and to SINDy in six systems. Performance gains prove particularly pronounced for systems exhibiting complex dynamics with trigonometric and high-order nonlinear terms (Thomas, Lorenz, and Aizawa systems). Critically, these improvements are accompanied by an approximate 100-fold computational speedup relative to ARGOS, establishing practical viability for large-scale applications.

Beyond performance metrics, the probabilistic formulation enables the deployment of standard statistical diagnostics, including PSIS-LOO~\cite{vehtariPracticalBayesianModel2017} for influential observations, VIF~\cite{marquardtRidgeRegressionPractice1975} for multicollinearity, and residual analysis for model misspecification, providing diagnostic signals for failure modes that are otherwise opaque. Importantly, the method's modular design facilitates integration with complementary identification pipelines for addressing higher-dimensional challenges. We demonstrate this extensibility by coupling Bayesian-ARGOS with the SINDy-SHRED~\cite{gaoSparseIdentificationNonlinear2025} framework for reduced-order modeling~\cite{bennerSurveyProjectionBasedModel2015} of sea surface temperature dynamics. Applied to neural-network-learned latent representations, this integration achieves a 77\% valid identification rate (compared to 60\% for standard SINDy) with markedly improved long-horizon prediction stability, effectively transforming a high-dimensional spatiotemporal problem into tractable latent dynamics identification. The discovered latent equations reveal a physically interpretable structure: a parsimonious affine-linear model capturing both the annual seasonal cycle (period $\approx 1.01$ years) and a fast transient mode (timescale $\approx 1.25$ years). Together, these advances show that automation, statistical rigor, and computational efficiency need not be traded against one another: governing equations can be reliably discovered from scarce and noisy observations, and the discovery process itself can be monitored to reveal when and why identification breaks down.

\section{Results}\label{sec2}
\subsection{Overview of the Bayesian-ARGOS Framework}

\begin{figure}[!htbp]
    \centering
    \includegraphics[width=\textwidth, trim=20pt 20pt 71pt 17pt,clip]{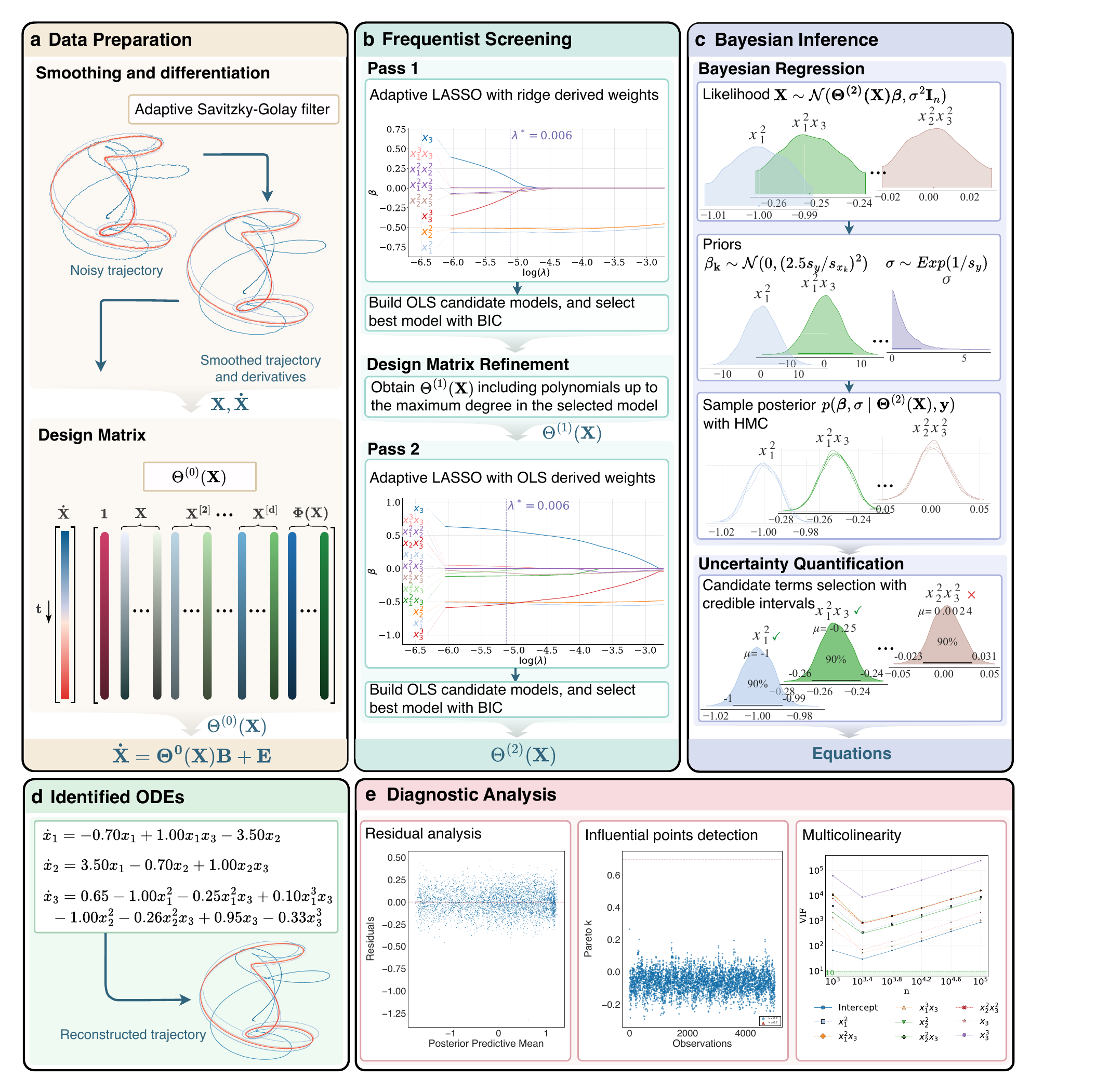}
    \caption{\textbf{Bayesian-ARGOS.} \textbf{a} Measured (noisy) state trajectories are smoothed and differentiated via an adaptive Savitzky--Golay filter to obtain $\mathbf{X}$ and $\dot{\mathbf{X}}$, from which a candidate library $\mathbf{\Theta}(\mathbf{X})$ is constructed. \textbf{b} A frequentist screening module performs a two-stage procedure---adaptive lasso with automated $\lambda$ selection, followed by threshold-based pruning with OLS refits and BIC model selection---in two sequential passes using Ridge- and OLS-derived adaptive weights, respectively. Between passes, the design matrix is refined to include terms up to the highest polynomial degree identified in the first pass, yielding a trimmed design matrix. \textbf{c} The Bayesian module applies HMC sampling to the trimmed design matrix to estimate posterior distributions of the coefficients and retain only terms whose credible intervals exclude zero. \textbf{d} and \textbf{e} The identified parsimonious ODEs are supplemented with statistical diagnostics---residual-structure analysis, influential-observation detection via PSIS-LOO,and multicollinearity assessment via VIF (applied before Bayesian regression)---as signals of potential model misspecification.}
    \label{fig:Bayesian_ARGOS_Schematic_Plot}
\end{figure}

Dynamical systems are commonly described by ordinary differential equations (ODEs):
\begin{equation}
    \dot{x}_j(t) = \frac{d}{dt}x_j(t) = f_j(x(t)), \qquad j = 1,\dotsc,m,
\end{equation}
where $x(t) = \left(x_{1}(t), x_{2}(t), \dotsc, x_{m}(t)\right)^T \in \mathbb{R}^{m}$ is the state vector and $f: \mathbb{R}^{m} \rightarrow \mathbb{R}^{m}$ governs the temporal evolution~\cite{guckenheimerNonlinearOscillationsDynamical1983}. Approximating each $f_j$ as a sparse linear combination of $p$ candidate basis functions transforms system identification into a regression problem:
\begin{equation}
    \dot{x}_j \approx \theta_{\text{F}}^T(x)\beta_j,  \qquad j = 1,\dotsc,m,
\end{equation}
where $\theta_{\text{F}}(x) \in \mathbb{R}^{p}$ is a feature vector of basis functions and $\beta_j \in \mathbb{R}^{p}$ is the corresponding sparse coefficient vector.

Given $n$ discrete measurements, we assemble the state matrix $\mathbf{X} \in \mathbb{R}^{n \times m}$, the derivative matrix $\dot{\mathbf{X}} \in \mathbb{R}^{n \times m}$ (obtained by measurement or numerical differentiation), and the design matrix $\mathbf{\Theta}(\mathbf{X})\in \mathbb{R}^{n\times p}$:
\begin{equation}
    \mathbf{\Theta}(\mathbf{X}) =
    \begin{pmatrix}
        \vert      & \vert      & \vert            &        & \vert            & \vert                     \\
        \mathbf{1} & \mathbf{X} & \mathbf{X}^{[2]} & \cdots & \mathbf{X}^{[d]} & \mathbf{\Phi}(\mathbf{X}) \\
        \vert      & \vert      & \vert            &        & \vert            & \vert
    \end{pmatrix}
\end{equation}
where $\mathbf{X}^{[i]}$ contains all degree-$i$ monomials and $\mathbf{\Phi}(\mathbf{X})$ includes additional nonlinear terms (e.g., trigonometric, logarithmic, exponential)~\cite{bruntonDiscoveringGoverningEquations2016}. The system identification problem then becomes:
\begin{equation}
    \dot{\mathbf{X}} = \mathbf{\Theta}(\mathbf{X})\mathbf{B} + \mathbf{E}
\end{equation}
where $\mathbf{B}\in \mathbb{R}^{p\times m}$ is the coefficient matrix and $\mathbf{E}\in\mathbb{R}^{n\times m}$ is the residual matrix. With an appropriate choice of basis functions in $\mathbf{\Theta}(\mathbf{X})$---whether monomials, trigonometric, or domain-specific---this formulation can represent the dynamics of systems ranging from chaotic oscillators to ecological interactions and gene regulatory circuits~\cite{bruntonDiscoveringGoverningEquations2016, kaptanogluBenchmarkingSparseSystem2023}. The central challenge is to identify the sparse coefficient matrix $\mathbf{B}$, that reveals which terms actively govern the evolution of each state variable $x_j(t)$.

The proposed Bayesian-ARGOS method solves for $\mathbf{B}$ via a hybrid sequential framework comprising a frequentist module for preliminary model identification followed by a Bayesian module for posterior inference and uncertainty quantification, as illustrated in Fig.~\ref{fig:Bayesian_ARGOS_Schematic_Plot} and described in Algorithm~\ref{al:Bayesian_ARGOS} in section~\ref{sec:algorithms} of the Supplementary Information.

The frequentist module applies a two-stage regression procedure executed in two passes. Within each pass, the adaptive lasso~\cite{zouAdaptiveLassoIts2006}, which augments the standard lasso~\cite{tibshiraniRegressionShrinkageSelection1996} with data-driven penalty weights, is first applied with an automatically tuned regularization parameter $\lambda$ to promote sparsity. Multiple candidate models are then generated by sweeping a sequence of thresholds over the resulting coefficients, retaining only terms whose magnitudes exceed each threshold. In the second stage, each candidate model is refitted by OLS regression to obtain unbiased coefficient estimates, and the optimal model is selected by minimizing the Bayesian Information Criterion (BIC)~\cite{schwarzEstimatingDimensionModel1978}. The two passes differ in their choice of adaptive weights: the first pass uses ridge-derived weights~\cite{hoerlRidgeRegressionBiased1970} to ensure robustness to multicollinearity, whereas the second uses OLS-derived weights to achieve asymptotic unbiasedness. Between passes, the design matrix is refined to include all functional terms up to the highest order of any variable with a nonzero coefficient identified in the first pass, thereby mitigating potential over-regularization~\cite{eganAutomaticallyDiscoveringOrdinary2024}. See Methods sections~\ref{sec:TwoStage} and~\ref{sec:sequential_execution_strategy} for details on the frequentist module.

The Bayesian module then operates on the refined design matrix produced by the frequentist module, employing Hamiltonian Monte Carlo (HMC)~\cite{betancourt2015hamiltonian} to draw samples from the posterior distribution of $\mathbf{B}$. The resulting posterior draws are used to construct marginal credible intervals for each coefficient, and a term is retained in the final model if its credible interval excludes zero, yielding an uncertainty-aware criterion for parsimonious model selection. See Methods section~\ref{sec:Bayesian_regression_and_uncertainty_quantification} for details on the Bayesian module.

\subsection{Assessing Bayesian-ARGOS on benchmark chaotic systems}
\label{subsection:Assessing_Bayesian-ARGOS_on_benchmark_chaotic_systems}

To assess the effectiveness of Bayesian-ARGOS, we selected seven three-dimensional chaotic systems that have been suggested as benchmarks for evaluating system identification methods recently~\cite{kaptanogluBenchmarkingSparseSystem2023}: the Lorenz, Thomas, Rössler, Dadras,  Aizawa, Sprott and Halvorsen  systems. For each system, we conducted a comprehensive evaluation by generating datasets under varying conditions. Specifically, we used 100 randomly selected initial conditions for each selected time series of length $n$ and signal-to-noise ratio (SNR). The random sampling of initial conditions from ranges that preserve the physical validity of each system is designed to simulate real-world scenarios where initial conditions are typically unknown a priori (see Supplementary Information section~\ref{supp_sect:benchmark_experimental_setup}). Meanwhile, the systematic variation of $n$ and SNR values enables comprehensive evaluation of algorithm performance across different data quantity and quality conditions~\cite{eganAutomaticallyDiscoveringOrdinary2024,liAutomatingDiscoveryPartial2024,gaoMeshfreeSparseIdentification2025}.

\begin{figure}[!htbp]
    \centering
    \includegraphics[width=\textwidth]{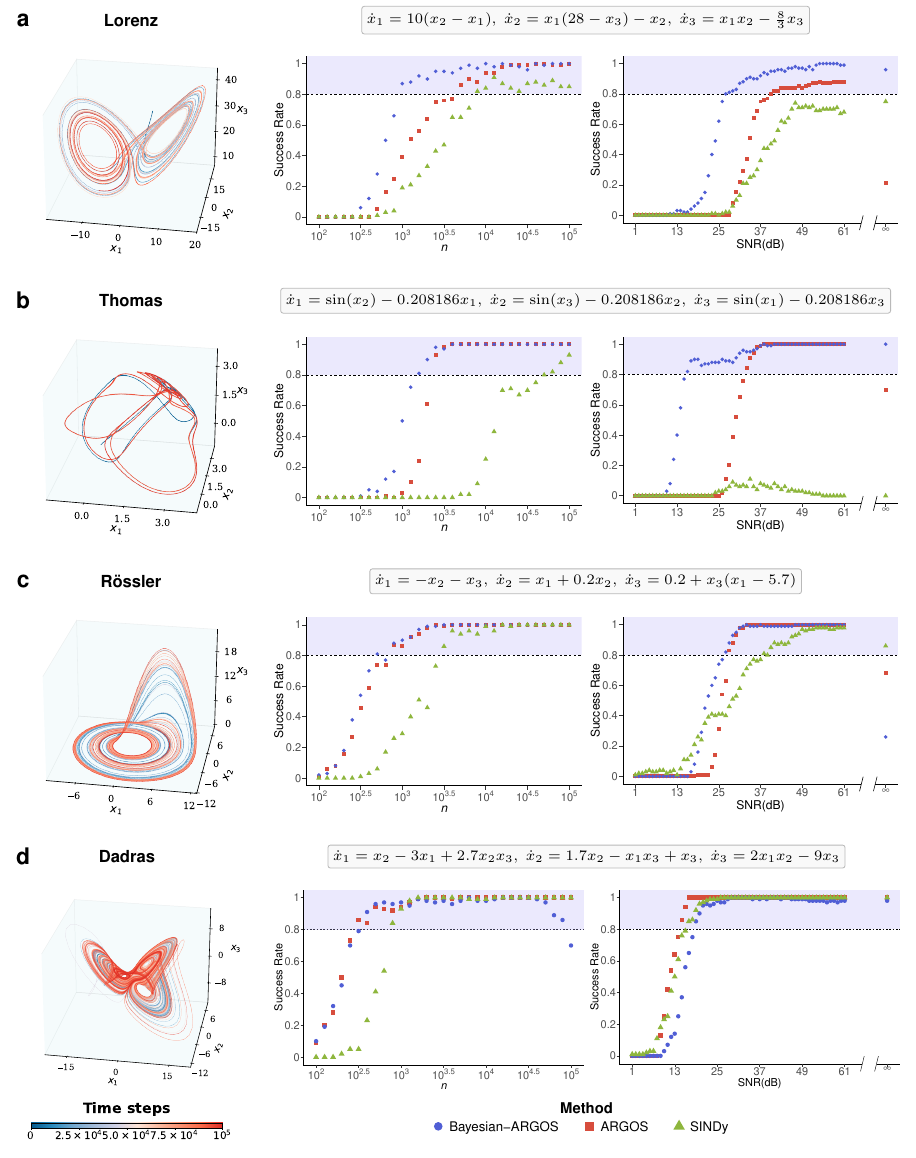}
    \caption{\textbf{Success-rate benchmarking for governing-equation identification on four chaotic systems.} \textbf{a}--\textbf{d} correspond to the Lorenz, Thomas, R\"ossler, and Dadras systems, respectively. For each system, the left subpanel shows a representative attractor; the middle subpanel shows the success rate (fraction of 100 trials recovering all ground-truth terms) versus number of observations $n$ at $\text{SNR} = 49~\text{dB}$; and the right subpanel shows the success rate versus SNR (dB) at $n=5000$. Success rates for Bayesian-ARGOS, ARGOS, and SINDy are shown as blue, red, and green, respectively. The dashed line marks the 80\% success-rate threshold.}
    \label{fig:Stacked_Identification_Results_1}
\end{figure}

\begin{figure}[!htbp]
    \centering
    \includegraphics[width=\textwidth]{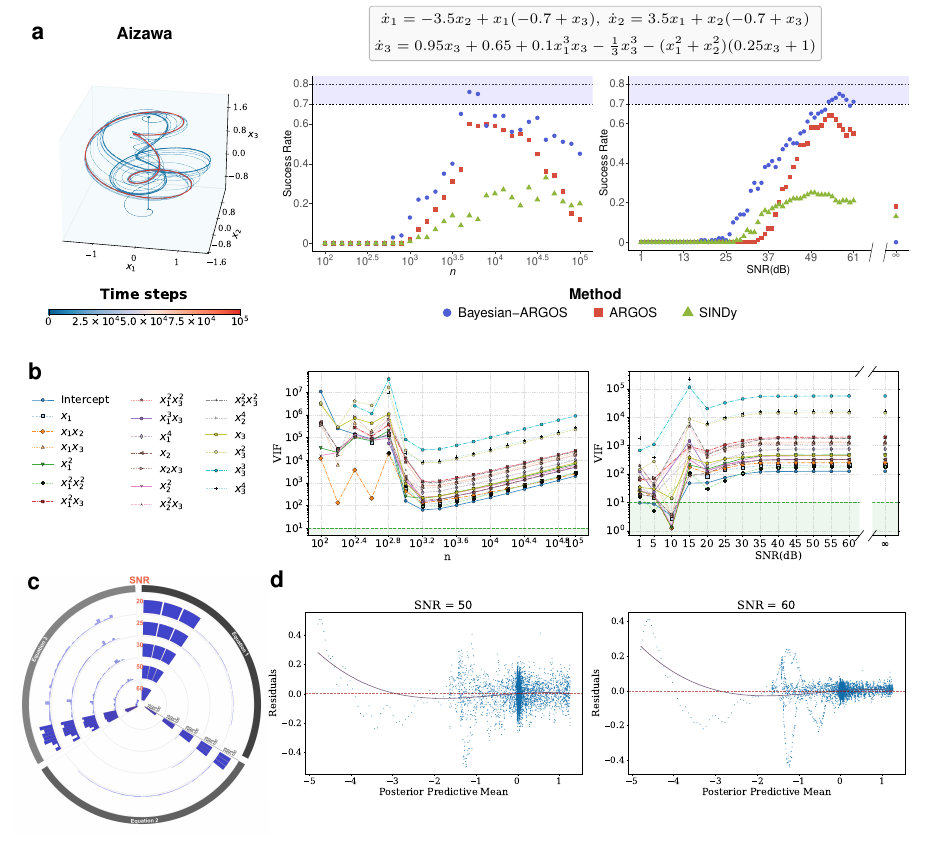}
    \caption{\textbf{Success-rate benchmarking and post-identification diagnostics for the Aizawa system.} \textbf{a} Representative attractor (left) and identification success rate (fraction of 100 trials that recover all ground-truth terms) versus the number of observations $n$ at fixed $\text{SNR} = 49~\text{dB}$ (middle) and versus SNR (dB) at fixed $n=5000$ (right) for Bayesian-ARGOS (blue), ARGOS (red), and SINDy (green); the dashed line marks the success-rate threshold (70\%). \textbf{b} Variance inflation factors (VIFs) for frequently selected candidate terms in $\dot{x}_3$ (terms listed in the left subpanel), shown across $n$ (middle) and SNR (right); the dashed line marks $\text{VIF}=10$, values above which indicate strong collinearity. \textbf{c} Identified-term distributions for the three equations at representative SNR values show spurious inclusion of extraneous terms in the $\dot{x}_3$ equation at $\text{SNR}=50~\text{dB}$ and $60~\text{dB}$. \textbf{d} Residuals versus posterior predictive mean for a representative $\dot{x}_3$ identification trial at $\text{SNR}=50~\text{dB}$ (left) and $60~\text{dB}$ (right) indicate the emergence of heteroscedastic structure at higher SNR.}
    \label{fig:Stacked_Identification_Results_2}
\end{figure}

To provide quantitative performance assessment, we employed the success rate metric, defined as the proportion of instances (out of 100 trials) where the algorithm correctly identified all governing equation terms for a given dynamical system at each specific $n$ and SNR values. We established an 80\% success rate threshold as the criterion for consistent and reliable identification of governing equations across most benchmark systems~\cite{eganAutomaticallyDiscoveringOrdinary2024}. For systems exhibiting higher nonlinearity, such as the Aizawa system, we reduced this threshold to 70\% to account for the increased complexity of the dynamics. Success rates exceeding these respective thresholds are highlighted. Across the seven benchmark systems, Bayesian-ARGOS demonstrates both data efficiency and noise robustness (Figs.~\ref{fig:Stacked_Identification_Results_1},~\ref{fig:Stacked_Identification_Results_2}a, and~\ref{fig:Stacked_Identification_Results_3}). At a fixed SNR of $49~\text{dB}$, five of seven systems reach the success-rate threshold with fewer than $10^{3.2}$ observations; three of these (the R\"ossler, Dadras, and Halvorsen systems) require only $10^{2.4}$--$10^{2.7}$ observations, indicating that even short time series suffice when the attractor geometry is well sampled. The Aizawa and Sprott systems are notable exceptions, requiring approximately $10^{3.7}$ observations. At a fixed sample size of $n=5000$, five systems surpass the 80\% threshold at an SNR of $27~\text{dB}$, while the Thomas and Halvorsen systems tolerate noise levels as low as $\text{SNR} = 17~\text{dB}$. Taken together, these results indicate that Bayesian-ARGOS reliably recovers governing equations from limited and noisy data, and highlight the system-dependent variability that motivates the diagnostic analysis in Section~\ref{section:Investigating the causes of anomalous performance degradation}.

\begin{figure*}[!htbp]
    \centering
    \includegraphics[width=\textwidth, trim=10pt 10pt 10pt 0pt,clip]{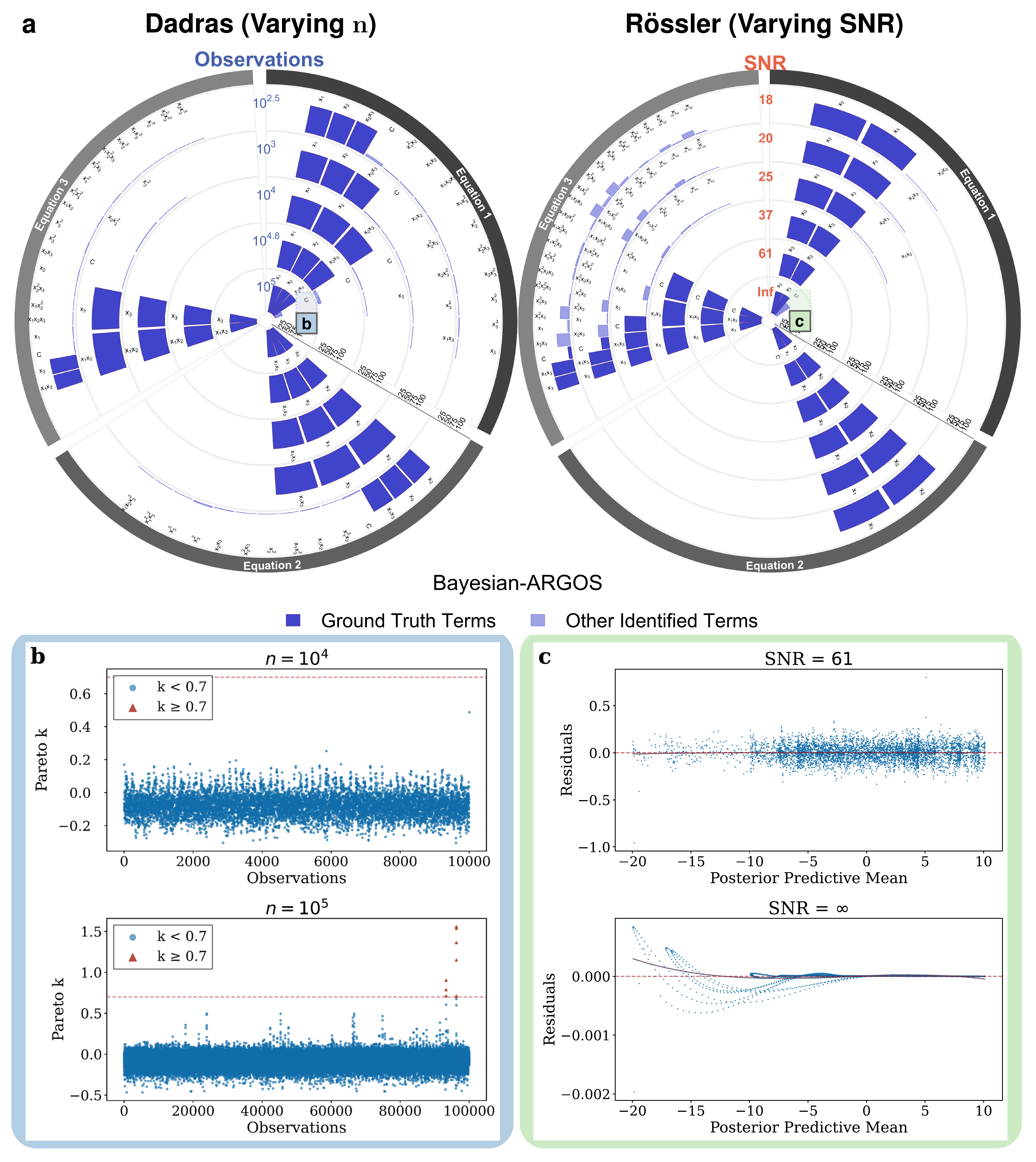}
    \caption{\textbf{Diagnostics of anomalous decreases in identification success rate at large $n$ and vanishing noise.} \textbf{a} Across 100 trials, term-selection frequencies for each governing equation of the three-dimensional systems in the regimes where performance degrades. Ground-truth terms are shown in dark blue and additionally selected terms in light blue. Spurious inclusion of an intercept term is highlighted and analyzed in the bottom panels (Dadras $\dot{x}_1$ at large $n$; R\"ossler $\dot{x}_1$ at $\text{SNR}=\infty$). \textbf{b} PSIS-LOO Pareto shape parameters $\hat{k}_i$ for a representative Dadras $\dot{x}_1$ trial, comparing $n=10^4$ (no influential points) and $n=10^5$ (many $\hat{k}_i>0.7$), indicating sensitivity to influential observations. \textbf{c} Residuals versus posterior predictive mean for the R\"ossler $\dot{x}_1$ equation at $\text{SNR}=\infty$ and $\text{SNR}=61~\text{dB}$, illustrating heteroscedastic structure at vanishing noise that can promote overselection.}
    \label{fig:diagnostic_plot_of_systems_with_drop_of_success_rate}
\end{figure*}

\subsection{Investigating the causes of anomalous performance degradation}
\label{section:Investigating the causes of anomalous performance degradation}

Assessment across the benchmark chaotic systems shows that Bayesian-ARGOS reliably recovers governing equations from limited and noisy data in most cases. Yet closer inspection reveals a counterintuitive phenomenon: in specific regimes, increasing data quantity or reducing noise degrades identification performance. This prompts us to investigate whether standard statistical diagnostics can identify the mechanisms underlying these anomalies.

For most systems, the success rate exhibits asymptotic behavior, approaching or reaching 100\% as the number of observations $n$ increases. Notable exceptions are the Aizawa and Dadras systems. The Aizawa system exhibits degradation in success rate when the sample size $n$ exceeds $10^{3.6}$. This failure is driven by severe multicollinearity within the design matrix, which we quantified using the Variance Inflation Factor (VIF)~\cite{marquardtRidgeRegressionPractice1975}. The VIF plot in Fig.~\ref{fig:Stacked_Identification_Results_2}b, computed from a single randomly selected identification trial at each $n$, shows that VIF values for frequently identified terms in the third equation ($\dot{x}_3$) of the Aizawa system consistently exceed the common problematic threshold of $10$ by several orders of magnitude~\cite{obrienCautionRegardingRules2007a}. Moreover, all VIF values for these terms increase by an order of magnitude as $n$ grows from $10^{3.4}$ to $10^5$. This extreme linear dependence indicates that these candidate features become nearly linearly dependent on the system's chaotic attractor manifold. Consequently, despite the availability of large datasets, the regression problem becomes ill-conditioned, preventing the algorithm from reliably distinguishing between these algebraically redundant terms and causing the observed drop in identification success rate.
%  The design matrix obtained from frequentist regression block

The Dadras system exhibits a decline in success rate when $n$ exceeds $10^{4.6}$ observations. To diagnose this anomaly, we employed Pareto-smoothed importance-sampling leave-one-out cross-validation (PSIS-LOO)~\cite{vehtariPracticalBayesianModel2017}, a Bayesian diagnostic tool that identifies influential observations by approximating leave-one-out cross-validation via importance sampling from the full posterior. The Pareto shape parameter, $\hat{k}_i$, serves as the key indicator; values exceeding $0.7$ signal observations that exert excessive influence on the posterior~\cite{vehtariParetoSmoothedImportance2024a}, potentially compromising the reliability of the approximation. Our analysis of a representative identification trial revealed that the dataset with $n=10^5$ observations contains substantially more influential points ($\hat{k}_i > 0.7$) compared to the $n=10^4$ case where no influential points are detected, as illustrated in Fig.~\ref{fig:diagnostic_plot_of_systems_with_drop_of_success_rate}b. These problematic points render the posterior inference susceptible to bias, clarifying the erratic term selection in the $\dot{x}_1$ equation: an extraneous intercept term is incorrectly incorporated due to the skewing effect of these high-influence observations, as indicated by the identified term frequencies of the Dadras system in Fig.~\ref{fig:diagnostic_plot_of_systems_with_drop_of_success_rate}a.

Across most benchmark systems, the success rate increases with SNR. Two anomalous regimes emerge at very low noise: the Aizawa system's success rate declines beyond $57~\text{dB}$, and performance deteriorates for the R\"ossler and Sprott systems at $\text{SNR}=\infty$. We first examine the Aizawa system. Unlike the degradation at large $n$, which arises from worsening multicollinearity, the SNR-related decline cannot be attributed to the same mechanism. VIF values for the $\dot{x}_3$ equation remain consistently high across all SNR levels (except at $\text{SNR}=10~\text{dB}$) but vary smoothly without an abrupt change near the threshold where the success rate decreases (Fig.~\ref{fig:Stacked_Identification_Results_2}b), indicating that multicollinearity alone does not account for this behavior. Instead, we identify heteroscedasticity as the primary cause. Figure~\ref{fig:Stacked_Identification_Results_2}d compares residuals versus posterior predictive means at $\text{SNR}=50~\text{dB}$ and $\text{SNR}=60~\text{dB}$. At $50~\text{dB}$, the residuals display a relatively diffuse, less structured pattern consistent with homoscedastic noise; at $60~\text{dB}$, they exhibit pronounced deterministic structure and mean-dependent variance, which are hallmarks of heteroscedasticity. This violation of the constant-variance assumption underlying the linear regression framework of Bayesian-ARGOS increases susceptibility to the inclusion of spurious terms: although ground-truth terms are recovered in most trials, additional spurious terms are incorrectly incorporated in $\dot{x}_3$, as shown by the identified-terms distribution in Fig.~\ref{fig:Stacked_Identification_Results_2}c.

The same heteroscedastic mechanism accounts for the performance deterioration of the R\"ossler and Sprott systems at $\text{SNR}=\infty$. The R\"ossler system's $\dot{x}_1$ equation provides a representative illustration. At $\text{SNR}=\infty$, the residuals versus posterior predictive mean plot exhibits a characteristic non-uniform spread indicative of heteroscedasticity (bottom panel of Fig.~\ref{fig:diagnostic_plot_of_systems_with_drop_of_success_rate}c). Because the homoscedastic Gaussian error model is misspecified under these conditions, uncertainty quantification becomes miscalibrated, and the algorithm adds spurious terms to the identified equations. The identified-terms distribution confirms that a spurious constant intercept is included in $\dot{x}_1$ (Fig.~\ref{fig:diagnostic_plot_of_systems_with_drop_of_success_rate}a). Remarkably, when noise increases only slightly from the noiseless condition to $\text{SNR}=61~\text{dB}$, the residual variance becomes more uniform, and Bayesian-ARGOS recovers its expected performance.

Taken together, these diagnostics indicate that the rare decreases in success rate stem from specific statistical pathologies rather than mere stochastic variability across trials. Extreme multicollinearity can render candidate terms nearly indistinguishable, making selection ill-conditioned as data volume grows (Aizawa at large $n$); influential observations can distort posterior-based model selection (Dadras at large $n$);  and near-noiseless data can expose misspecification of the homoscedastic Gaussian error model and promote over-selection of spurious terms (R\"ossler at $\text{SNR}=\infty$ and Aizawa at high SNR). Two broader implications follow. First, ``more data'' and ``less noise'' are not uniformly beneficial for regression-based chaotic system identification: dense sampling on a low-dimensional attractor can amplify feature redundancy, while vanishing noise can reveal deterministic residual structures. Second, these failure modes are diagnosable and therefore actionable. By leveraging influence diagnostics with PSIS-LOO, multicollinearity checks with VIF, and residual checks, we can delineate the precise regimes where Bayesian-ARGOS remains reliably data-efficient and noise-robust.
%  Post identification diagnostics

\subsection{Comparisons with ARGOS and SINDy}

To conduct a comprehensive comparative evaluation, we compared Bayesian-ARGOS with the two baseline methods  ARGOS~\cite{eganAutomaticallyDiscoveringOrdinary2024}  and SINDy~\cite{bruntonDiscoveringGoverningEquations2016} across the seven selected benchmark chaotic systems, see section~\ref{sec: performance_comparison} of Methods for details. To ensure a fair comparison, all methods incorporated the same adaptive Savitzky-Golay filtering approach~\cite{eganAutomaticallyDiscoveringOrdinary2024} and utilized a candidate library containing polynomial terms up to the fifth order, totaling 56 candidate terms including an intercept. For the Thomas system containing trigonometric components, the candidate library was augmented with first-order sine and cosine terms, expanding to 62 total candidate terms.

\begin{figure*}[!htbp]
    \centering
    \includegraphics[width=1\textwidth]{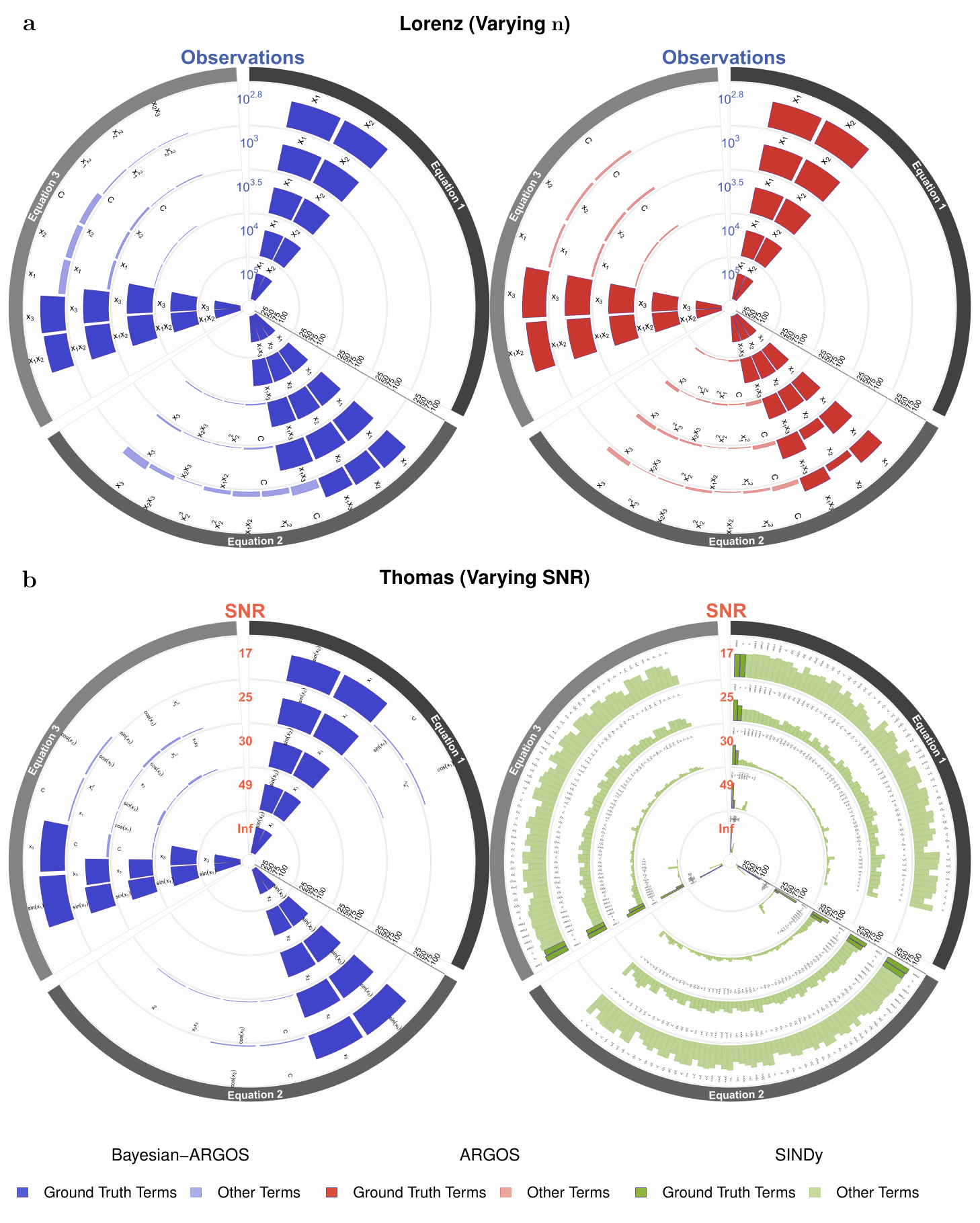}
    \caption{\textbf{Shrinkage behavior of Bayesian-ARGOS versus ARGOS and SINDy.} \textbf{a} Lorenz system: term-selection frequencies for Bayesian-ARGOS (left) and ARGOS (right) as the number of observations ($n$) varies (concentric rings), illustrating that ARGOS can over-shrink and omit the ground-truth $x_2$ term in $\dot{x}_2$ at moderate $n$. \textbf{b} Thomas system: selection frequencies for Bayesian-ARGOS (left) and SINDy (right) across SNR (concentric rings; \textbf{Inf} denotes noiseless data), showing that Bayesian-ARGOS concentrates selection on ground-truth terms whereas SINDy more frequently includes spurious terms under noise. Outer grey sectors indicate equations; bars correspond to identified library terms and bar height indicates the fraction of independent runs in which the term is selected. Dark colors denote ground-truth terms and light colors denote other selected terms.}
    \label{fig:Cumulated Distribution of Identified Terms}
\end{figure*}

When varying the number of observations, $n$, Bayesian-ARGOS reaches the corresponding success-rate thresholds with fewer observations than ARGOS in five of the seven systems (see Figs.~\ref{fig:Stacked_Identification_Results_1},~\ref{fig:Stacked_Identification_Results_2}a, and~\ref{fig:Stacked_Identification_Results_3}). The Dadras and Sprott systems are exceptions, for which the differences between the two methods remain minimal. Compared with SINDy, Bayesian-ARGOS demonstrates consistently faster performance improvement as $n$ increases across all systems. The largest advantage of Bayesian-ARGOS appears at moderate sample sizes, ranging from $10^{2.5}$ to $10^{3}$ observations, for the Lorenz and Thomas systems, the latter of which includes trigonometric components. To understand the mechanism behind this improvement, we compared the distributions of identified terms for the Lorenz system produced by Bayesian-ARGOS and ARGOS. As shown in Fig.~\ref{fig:Cumulated Distribution of Identified Terms}a, Bayesian-ARGOS more frequently recovers the $x_2$ term in the $\dot{x}_2$ equation of the Lorenz system, whereas ARGOS often omits this term within the $10^{2.5}$ to $10^{3}$ observation range. This omission arises from the adaptive lasso's aggressive shrinkage combined with the cross-validation procedure in the ARGOS algorithm, which can over-penalize ground-truth terms. Notably, for the Aizawa system, which is characterized by extreme multicollinearity due to multiple higher-order terms in its third equation, Bayesian-ARGOS achieves superior success rates across all values of $n$ compared with both baselines, demonstrating exceptional robustness to ill-conditioned design matrices, as discussed in Section~\ref{section:Investigating the causes of anomalous performance degradation} (see Fig.~\ref{fig:Stacked_Identification_Results_2}a).

Under varying SNR conditions, Bayesian-ARGOS demonstrates enhanced noise robustness, reaching the corresponding success rate thresholds at lower SNR than ARGOS in four of seven systems and outperforming SINDy in all systems except the Dadras system. This advantage is most pronounced for the Lorenz, Thomas, and Aizawa systems, where Bayesian-ARGOS matches or exceeds both baselines despite complex dynamics and strong noise perturbations. The contrast with SINDy is particularly stark for the Thomas and Aizawa systems: while SINDy exhibits success rates persistently below 30\% across SNR values, Bayesian-ARGOS achieves substantially higher performance. This systematic difference stems from a fundamental algorithmic distinction. As shown in Fig.~\ref{fig:Cumulated Distribution of Identified Terms}b, Bayesian-ARGOS concentrates selection frequency on ground-truth terms across noise levels, whereas SINDy exhibits broader spread and more frequently includes spurious terms. Mechanistically, Bayesian-ARGOS employs regularized selection with conditional uncertainty quantification that systematically penalizes noise induced overselection, whereas SINDy's thresholding mechanism with manually selected parameters has limited ability to distinguish signal driven from noise driven terms when the library becomes correlated under noise. This enhanced robustness, however, reflects a deliberate regularization trade-off rather than unqualified superiority. Bayesian-ARGOS employs balanced shrinkage calibrated to suppress most false positives while avoiding overly aggressive sparsification that could induce term-specific false negatives under severe noise. This design choice explains both the method's general robustness and its modest underperformance in specific regimes. For the Sprott and Halvorsen systems, Bayesian-ARGOS exhibits marginally weaker performance than ARGOS, though differences remain small. More notably, for the Dadras system at low SNR ($10$ to $20$~dB), Bayesian-ARGOS exhibits modestly broader term selection with slightly lower success rates compared to both baselines. Critically, this behavior is not a failure but rather the predictable consequence of the shrinkage calibration: by avoiding excessive penalty on ground-truth terms in challenging conditions, the method occasionally admits additional spurious terms when noise overwhelms the signal (see supplementary information~\ref{supp_sect:benchmark_experimental_setup}). This pattern reveals a fundamental tension in sparse identification under severe noise: the optimal balance between false-positive suppression and false-negative avoidance is inherently system-dependent, so a shrinkage calibration that works well broadly may underperform in specific regimes.

\begin{figure*}[!htbp]
    \centering
    \includegraphics[width=\textwidth]{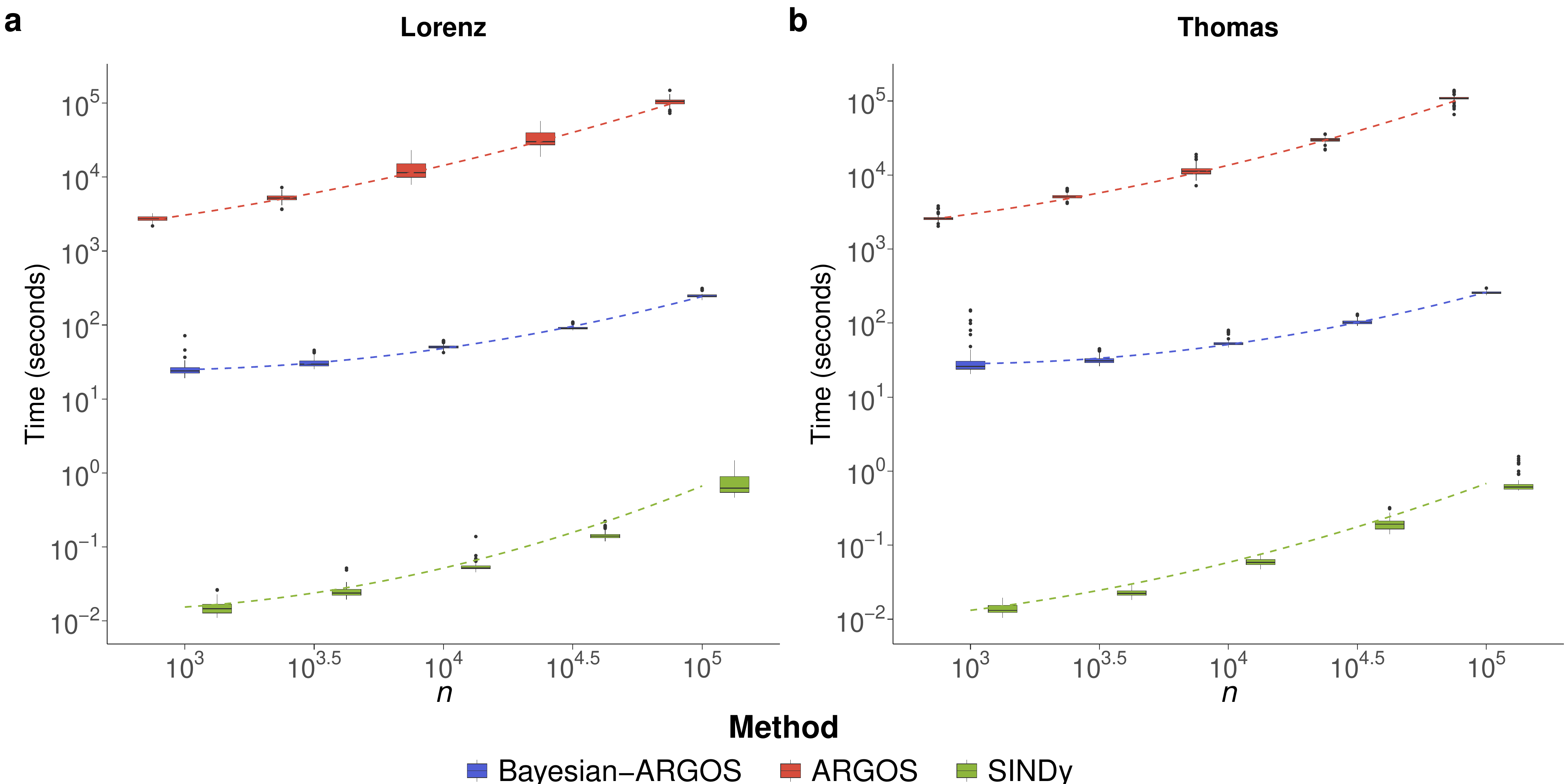}
    \caption{\textbf{Computational runtime scaling.} Runtime (seconds) of Bayesian-ARGOS, ARGOS, and SINDy as a function of the number of observations ($n$) ($10^{3}$ to $10^{5}$). Bayesian-ARGOS achieves an approximately two-orders-of-magnitude speedup over ARGOS;
        %  As $n$ increases, the computational advantage of Bayesian-ARGOS over ARGOS continues to grow.
        SINDy remains the fastest baseline but does not provide uncertainty quantification. In each boxplot, the black line denotes the median; whiskers extend to 1.5 times the interquartile range; points beyond the whiskers are outliers. Dashed lines indicate the fitted mean runtime as a function of $n$ for each algorithm.}
    \label{fig:Computational_Cost}
\end{figure*}

Accuracy alone is not sufficient for large-scale dynamical discovery, computational efficiency is also an important factor. As shown in Fig.~\ref{fig:Computational_Cost}, Bayesian-ARGOS delivers a two order of magnitude acceleration compared to standard ARGOS across dataset sizes $n=10^{3}$ to $10^{5}$. Specifically, at $n=10^{5}$, Bayesian-ARGOS reduces execution time from $>10^{4.7}$ seconds (ARGOS) to $<10^{2.5}$ seconds. Although SINDy remains the fastest method due to its deterministic, fixed-iteration architecture, this speed comes at the cost of robustness and the omission of uncertainty quantification. Furthermore, Bayesian-ARGOS demonstrates superior consistency, with lower computational variability across trials. By maintaining practical execution times without sacrificing principled uncertainty estimation, Bayesian-ARGOS effectively bridges the gap between the speed of SINDy and the statistical rigor of ARGOS.

Collectively, these comparative results establish Bayesian-ARGOS as a robust framework for governing equation identification that strategically balances statistical rigor, identification accuracy, and computational efficiency. First, the method achieves superior data efficiency compared to both baselines, requiring fewer observations to reach reliable identification thresholds in the majority of benchmark systems. Second, it demonstrates enhanced noise robustness. Third, these performance gains are achieved without compromising computational tractability: the two-orders-of-magnitude speedup relative to ARGOS positions Bayesian-ARGOS as practically viable for large-scale applications while preserving uncertainty quantification that is absent in deterministic methods like SINDy.
% The method's particular effectiveness for systems with extreme multicollinearity (e.g., Aizawa) and complex trigonometric dynamics (e.g., Thomas) underscores the value of its hybrid architecture. Specifically, the frequentist adaptive lasso component efficiently narrows the model space to mitigate computational burden, while the Bayesian inference component provides principled regularization that stabilizes selection when design matrices become ill-conditioned under noise or feature correlation.
However, the marginal performance degradation observed for certain systems (Sprott, Halvorsen) under specific noise conditions indicates that no single method universally dominates across all scenarios, reinforcing the importance of diagnostic-guided method selection tailored to specific system characteristics and data conditions.

\subsection{Integration with SINDy-SHRED in dealing with High-Dimensional problem}

The benchmark results above demonstrate that Bayesian-ARGOS
has strong performance on identifying classical dynamical systems. A more ambitious test asks whether the framework can identify interpretable dynamics in systems where the true equations are unknown and the observations are
high-dimensional.

Real-world spatiotemporal data presents substantial challenges including high dimensionality, measurement noise, and limited observations~\cite{xuRobustDiscoveryPartial2021, luDiscoveringSparseInterpretable2022, duDISCOVERDeepIdentification2024}, suggesting that standalone library-based sparse learning methods may prove insufficient for practical applications~\cite{championDatadrivenDiscoveryCoordinates2019,chenPhysicsinformedLearningGoverning2021, reinboldRobustLearningNoisy2021, gaoDatadrivenInferenceComplex2023,wuDatadrivenModelDiscovery2025}. Recent advances demonstrate that integrating neural networks with sparse regression methods substantially enhances library-based learning by synergistically combining the interpretability of sparse regression with the representational capacity of neural networks to capture complex spatiotemporal patterns~\cite{championDatadrivenDiscoveryCoordinates2019, gaoLearningInterpretableDynamics2024, gaoSparseIdentificationNonlinear2025}. Such integration has achieved success in identifying insightful latent dynamics in neural activities \cite{rautArousalUniversalEmbedding2025}. Therefore, the integration of Bayesian-ARGOS with neural network architectures presents a promising solution for complex real-world problems that exceed traditional sparse learning capabilities.

\begin{figure}[!htbp]
    \centering
    \includegraphics[width=\textwidth, trim=21pt 28pt 21pt 14pt,clip]{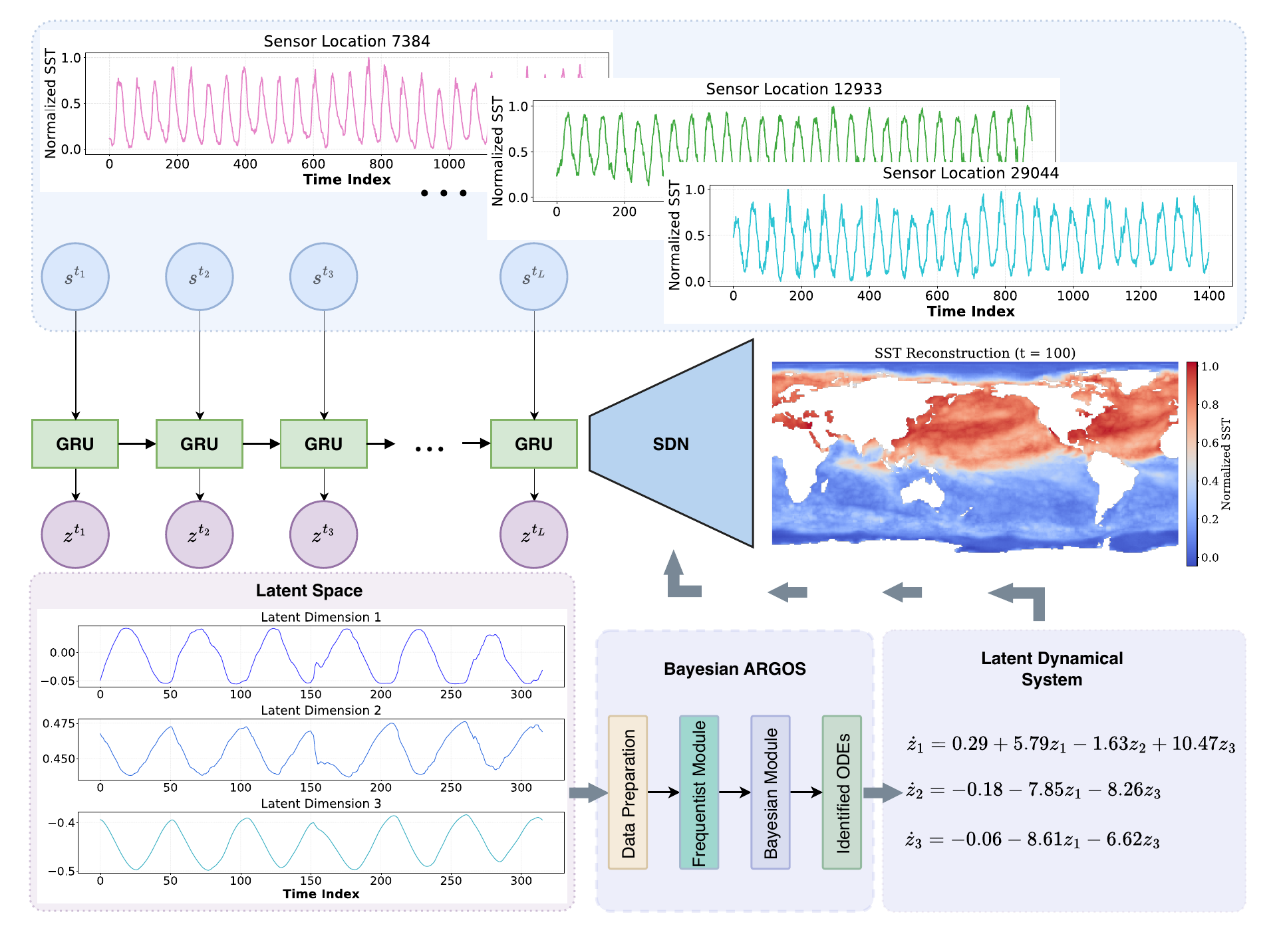}
    \caption{\textbf{Integration of Bayesian-ARGOS with SINDy-SHRED.} Sparse sensor time series are mapped by a sequence model into a low-dimensional latent state; a shallow decoder reconstructs the full spatiotemporal field, while a latent forecaster identifies governing equations and advances the latent dynamics. Bayesian-ARGOS here functions as the latent forecaster to provide posterior distributions over latent-dynamics coefficients and uncertainty-aware forecasts that are subsequently decoded to produce field predictions.}
    \label{fig:Methods_Integration_Schematic_Plot}
\end{figure}

To demonstrate Bayesian-ARGOS's effectiveness when combined with neural networks and its applicability to real-world problems, we integrate it into the SINDy-SHRED (Sparse Identification of Nonlinear Dynamics with Shallow Recurrent Decoder networks) framework~\cite{gaoSparseIdentificationNonlinear2025} for spatiotemporal identification using sea surface temperature sensor measurements. SINDy-SHRED reconstructs and forecasts high-dimensional spatiotemporal systems from sparse sensor measurements via three components: a sequence model encoding temporal measurements into latent space, a shallow decoder reconstructing the full spatiotemporal field~\cite{tomasettoReducedOrderModeling2025a}, and a latent forecaster identifying governing equations in the latent space. Together, the sequence model and decoder establish latent coordinates that enable parsimonious equation discovery. The latent forecaster is critical, as its performance determines the validity of discovered equations and long-term prediction accuracy. We replace SINDy, the original latent forecaster, with Bayesian-ARGOS and compare their performance. The integrated pipeline is illustrated in Fig.~\ref{fig:Methods_Integration_Schematic_Plot}.

We apply the integrated framework to global sea surface temperature (SST) data from the National Oceanic and Atmospheric Administration (NOAA)~\cite{reynoldsImprovedSituSatellite2002a}, comprising $1,400$ weekly measurements from 1992 to 2019 on a $180 \times 360$ global grid. Of the total grid points, $44,219$ correspond to ocean surface locations. We apply min-max normalization to scale SST values at each grid point to the range $(0, 1)$.

We conducted $150$ independent training experiments of the SINDy-SHRED network, each using $250$ randomly selected sensors from $44,219$ ocean locations. Each training sample consisted of $52$-week SST time series from the selected sensors as input and the complete SST field across all grid locations at the subsequent time step as output. These $150$ experiments generated $150$ distinct three-dimensional latent spaces, each providing a unique lower-dimensional representation of the underlying SST dynamics. We implemented quality control by filtering out cases where the relative $\ell_2$ error (rRMSE) between reconstructed and ground truth SST values on the test dataset exceeded $0.05$, retaining only latent spaces with sufficient representational capacity for robust system identification.

After the quality filtering described above, $107$ high-quality latent representations remained for governing equation identification. We configured both Bayesian-ARGOS and SINDy as latent forecasters using candidate libraries with first-order polynomial terms, and set the SINDy hard threshold to $0.8$~\cite{gaoSparseIdentificationNonlinear2025}. We then compared the two methods using two metrics. The first was the mean squared error (MSE) between latent-space trajectories simulated from the identified equations and the corresponding ground truth trajectories derived from the trained sequence model. The second was the root mean squared error (RMSE) between the reconstructed spatiotemporal fields, obtained by decoding the simulated trajectories back into the high-dimensional space, and the corresponding ground truth SST fields.

\begin{figure*}[!htbp]
    \centering
    \begin{minipage}[c]{0.50 \linewidth}
        \includegraphics[width=\linewidth]{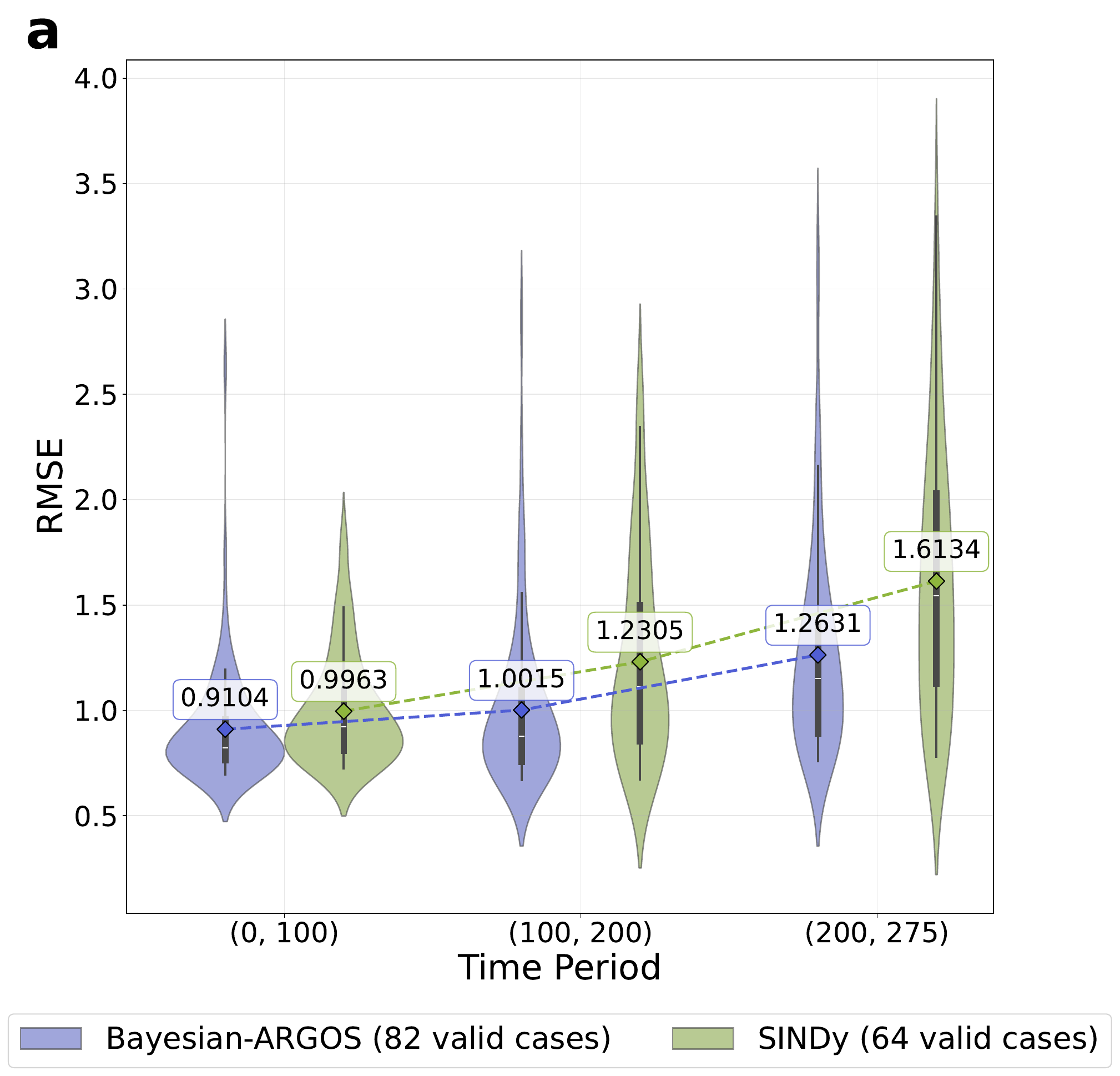}
    \end{minipage}
    \begin{minipage}[c]{0.485\linewidth}
        % \vspace{1pt}
        \includegraphics[width=\linewidth]{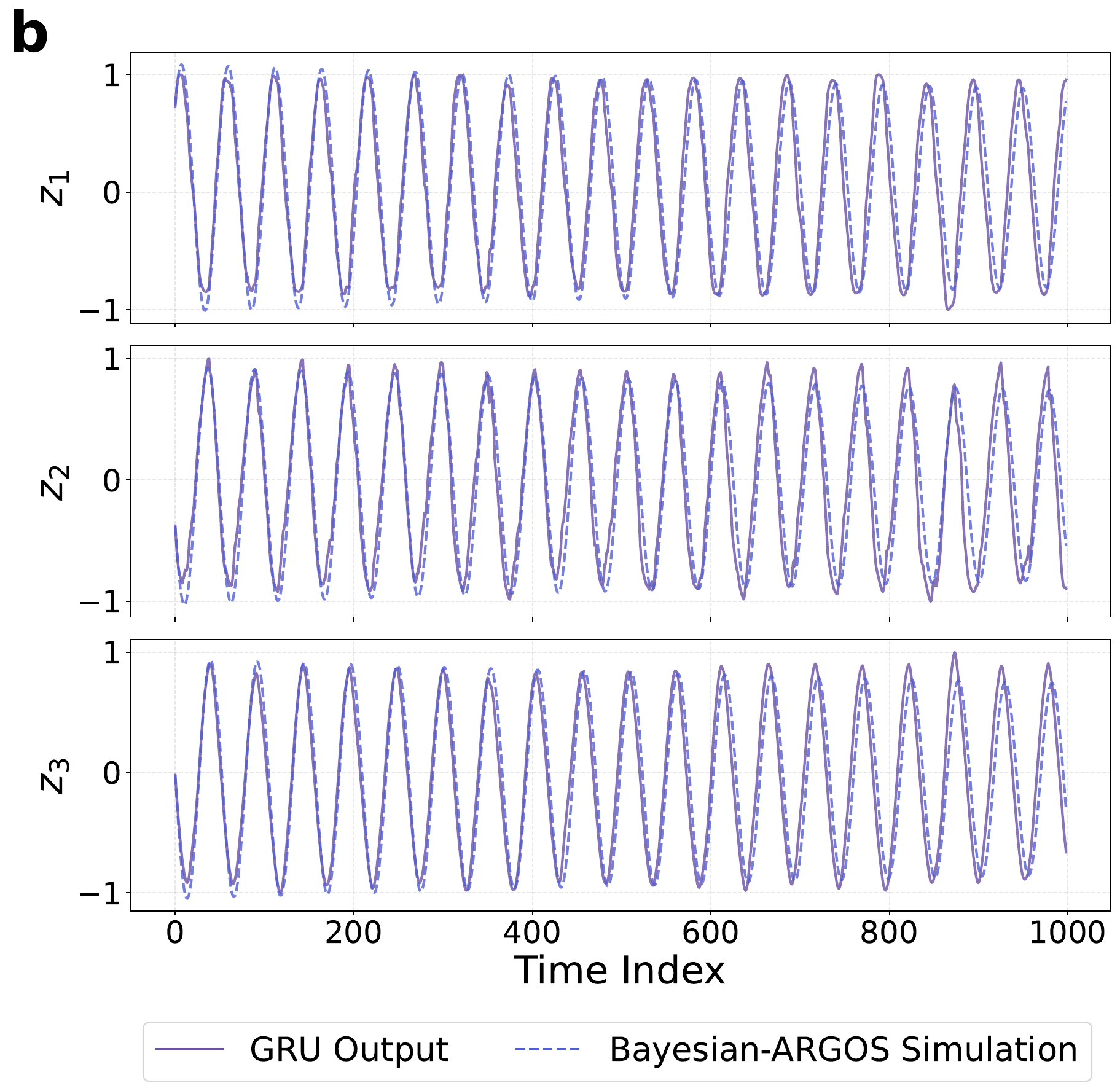}
    \end{minipage}
    \vspace{5pt}
    \hspace*{5pt} \includegraphics[width=\linewidth]{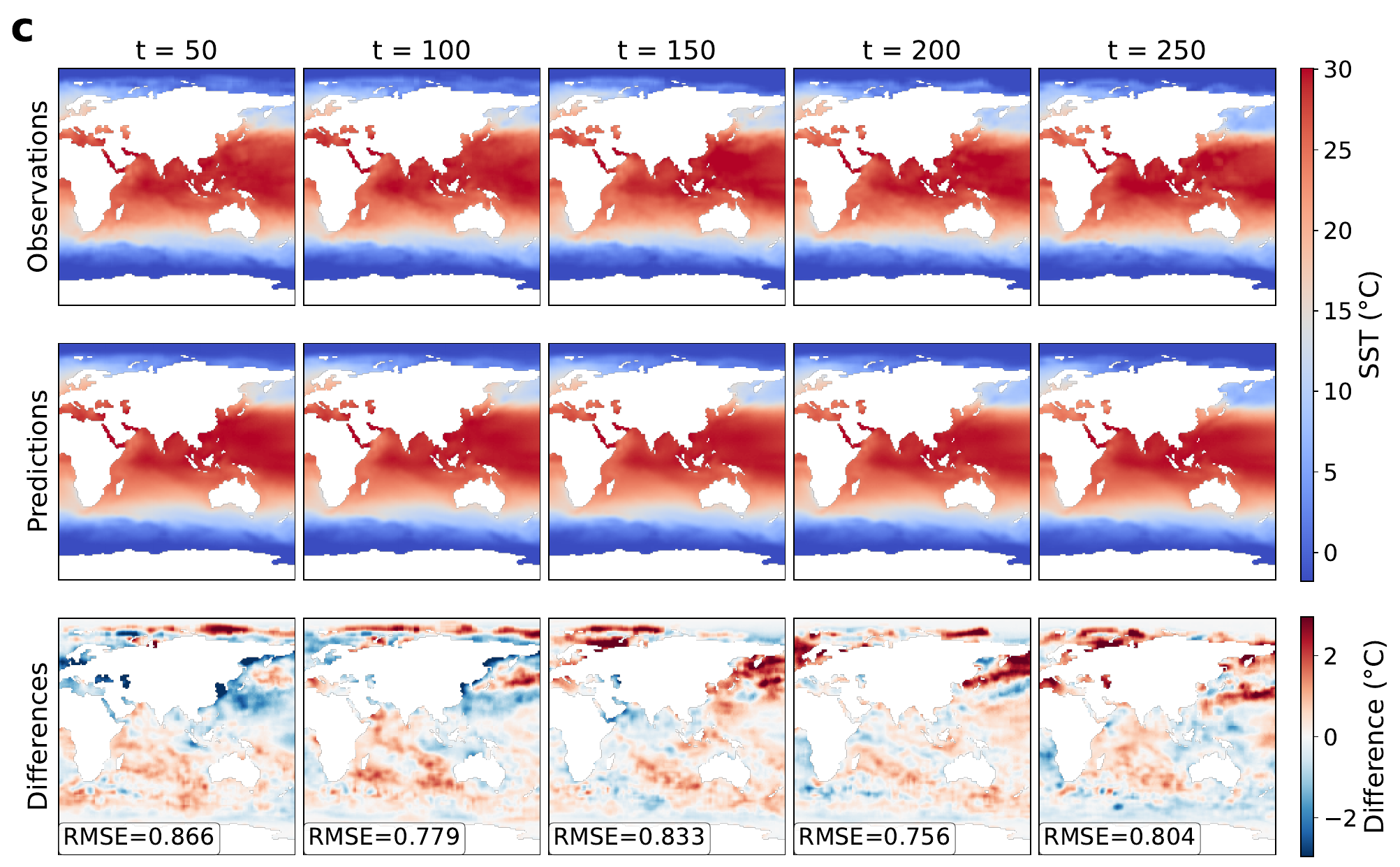}
    \caption{\textbf{Integrating Bayesian-ARGOS with SINDy-SHRED improves the robustness of latent-dynamics identification and long-horizon forecasting for global sea surface temperature (SST).} \textbf{a} Distribution of decoded-field reconstruction RMSE across learned latent spaces for three forecast intervals, comparing Bayesian-ARGOS (blue) with the original SINDy forecaster (green); diamonds indicate means and box overlays indicate interquartile ranges. \textbf{b} Representative trajectories in a learned 3D latent state ($z=(z_1,z_2,z_3)^\top$): GRU-derived latent evolution (solid) and Bayesian-ARGOS simulations from the identified governing equations (dashed). \textbf{c} Decoded SST forecasts at increasing lead times (from $50$ to $250$ time steps): observations (top), Bayesian-ARGOS predictions (middle) and differences (bottom); RMSE is reported for each horizon, and colour bars indicate SST ($^\circ\text{C}$) and prediction error($^\circ\text{C}$).}
    \label{fig:SINDy_SHRED_Integration_Results}
\end{figure*}

To ensure the identified equations provided a reliable foundation for spatiotemporal reconstruction, we imposed an additional validity filter in the latent space. Specifically, we discarded a case if the latent-trajectory Mean Squared Error (MSE) in any of the three latent dimensions exceeded a threshold of $1.0$, which is calibrated to the normalization scale of the SST values; a large error in a single coordinate is sufficient to corrupt decoding and indicates substantial information loss. This criterion revealed a marked difference in robustness: Bayesian-ARGOS produced valid governing equations in $77\%$ of cases ($82/107$), whereas SINDy succeeded in $60\%$ ($64/107$). Among the valid cases, Bayesian-ARGOS also achieved higher accuracy,  with an average latent MSE of $0.263$ and an average reconstruction RMSE of $1.055$, compared with $0.334$ and $1.282$ for SINDy, respectively. Notably, the performance gap widens for longer-horizon forecasts.  As shown in Fig.~\ref{fig:SINDy_SHRED_Integration_Results}a, Bayesian-ARGOS integration yields consistently lower reconstruction RMSE across all three prediction intervals, and exhibits substantially reduced degradation for longer-terms predictions compared to SINDy.

We highlight a representative case in which Bayesian-ARGOS identifies the following parsimonious affine-linear dynamics in a learned three-dimensional latent state $z = (z_1, z_2, z_3)^\top$:
\begin{equation}
    \begin{cases}
        \dot{z}_1 = \phantom{-}0.29 + 5.79z_1 - 1.63z_2 + 10.47z_3, \\
        \dot{z}_2 = -0.18 - 7.85z_1 - 8.26z_3,                      \\
        \dot{z}_3 = -0.06 - 8.61z_1 - 6.62z_3.
    \end{cases}
    \label{eq:latent_dynamics_identified}
\end{equation}
Writing Eq.~\eqref{eq:latent_dynamics_identified} compactly as $\dot{z} = Az + b$ implies an equilibrium at $z^* = -A^{-1}b \approx (0.036, -0.055, -0.056)^\top$. Centering about this equilibrium via $y(t) = z(t) - z^*$ removes the constant offset and yields the homogeneous linear system $\dot{y} = Ay$. This mean-shifted linear drift is qualitatively consistent with the Linear Inverse Modeling / Principal Oscillation Pattern (LIM/POP) viewpoint for SST anomalies, where unresolved processes are often represented as stochastic forcing and oscillations arise from complex-conjugate eigenpairs of the drift matrix~\cite{penlandRandomForcingForecasting1989, penlandStochasticModelIndoPacific1996, alexanderForecastingPacificSSTs2008}.

Eigenanalysis of $A$ yields a complex-conjugate pair $\lambda_{1,2} \approx -0.015 \pm 6.24i$ and a real eigenvalue $\lambda_3 \approx -0.80$, so the solution decomposes into a fast decaying mode and a weakly damped oscillator:
\begin{equation}
    y(t) = c_1 v_{\mathrm{fast}} e^{-0.80 t} + e^{-0.015 t}\Big( c_2 v_{\mathrm{c}}\cos(6.24 t) + c_3 v_{\mathrm{s}}\sin(6.24 t)\Big),
    \label{eq:analytical_solution_of_latent_dynamics}
\end{equation}
The oscillatory mode has angular frequency $\omega \approx 6.24~\mathrm{rad\,yr^{-1}}$ (period $T = 2\pi/\omega \approx 1.01$ years) and a small decay rate of $0.015~\mathrm{yr^{-1}}$ (half-life $\ln 2/0.015 \approx 46$ years), while the real mode decays on a timescale $1/0.80 \approx 1.25$ years. Consistent with this stable modal structure, simulated latent trajectories (dashed) closely track the GRU-derived ground truth (solid) with minimal phase drift (Fig.~\ref{fig:SINDy_SHRED_Integration_Results}b). Decoding these latent simulations through the shallow decoder yields stable and accurate spatiotemporal forecasts across horizons from 50 to 250 weeks (in increments of 50), without abrupt divergence as errors accumulate (Fig.~\ref{fig:SINDy_SHRED_Integration_Results}c). Consistent behavior is also observed at individual sensor locations, where decoded forecasts provide stable extrapolations across randomly sampled test sensors (Appendix Fig.~\ref{fig:multi_sensor_forecast}).

% The identified dynamics exhibit three distinct timescales. The complex conjugate eigenvalues $\lambda_{1,2} \approx -0.015 \pm 6.24i$ generate an oscillatory mode with frequency $\omega \approx 6.24~\mathrm{rad\,yr^{-1}}$ (period $T \approx 1.01$ years), capturing the annual seasonal cycle in SST. This oscillation decays slowly with rate $0.015~\mathrm{yr^{-1}}$ (half-life $\approx 46$ years), ensuring long-term stability while preserving the seasonal signal. The real eigenvalue $\lambda_3 \approx -0.80$ corresponds to a fast transient mode with timescale $\tau \approx 1.25$ years that rapidly dissipates initial perturbations. Figure~\ref{fig:SINDy_SHRED_Integration_Results}b demonstrates that simulated latent trajectories (dashed) closely match the ground truth (solid) with minimal phase drift, validating the identified model's accuracy in the latent space.

Integrating Bayesian-ARGOS as the latent forecasting module provides two central advantages for high-dimensional system identification. First, it substantially improves robustness and long-horizon stability: across hundreds of randomly sampled sensor configurations, Bayesian-ARGOS more reliably identifies latent dynamics and delivers reconstructions whose errors grow more slowly with forecast horizon. Second, it replaces ad hoc sparsification choices with a principled treatment of sparsity and uncertainty, reducing sensitivity to hyperparameters. By coupling uncertainty-aware sparse regression with neural embeddings and decoding, the integration of Bayesian-ARGOS into SINDy-SHRED framework offers a practical route to interpretable governing equations and stable spatiotemporal predictions in realistic, noisy, multi-scale systems.

\section{Discussion}\label{sec:Discussion}

The results presented here demonstrate that the competing demands of automation, statistical rigor, and computational efficiency in equation discovery can be reconciled through a principled division of labor: frequentist screening rapidly narrows the candidate space to a parsimonious subset, while Bayesian inference provides rigorous uncertainty quantification over the resulting reduced model. This complementarity, rather than any single algorithmic innovation, constitutes the central contribution of the Bayesian-ARGOS framework.
% support statistical diagnostics

A particularly significant feature of this design is the diagnostic transparency it confers. Contrary to naive expectation, increasing data quantity or reducing noise does not guarantee improved identification: influential observations can distort posterior inference~\cite{vehtariPracticalBayesianModel2017}, dense sampling on attractors can amplify multicollinearity~\cite{belsley2005regression}, and vanishing noise can expose violations of the homoscedastic error assumption~\cite{gelmanPosteriorPredictiveAssessment1996}. These pathological regimes, which remain opaque in purely algorithmic implementations, become detectable and diagnosable through standard statistical tools---PSIS-LOO influence metrics, VIF-based multicollinearity detection, and posterior predictive residual analysis. The framework thereby serves a dual role: as a sparse identification engine and as a reliability monitor that converts opaque failures into actionable diagnoses, pointing toward targeted remedies such as the removal of influential observations~\cite{cookResidualsInfluenceRegression1982}, collection of data from undersampled attractor regions, or refinement of the candidate library to reduce inter-term correlation~\cite{boninsegnaSparseLearningStochastic2018a}.

Integrating Bayesian-ARGOS into the SINDy-SHRED pipeline demonstrates the framework's broader potential as a modular component within high-dimensional discovery workflows. Replacing deterministic sparsification with uncertainty-aware inference in the latent space increases the fraction of identifiable governing equations and improves long-horizon prediction stability~\cite{bengioRepresentationLearningReview2013}. The resulting workflow---neural dimensionality reduction followed by uncertainty-aware equation discovery---provides a scalable template for inferring governing equations from high-dimensional spatiotemporal observations. The modularity of Bayesian-ARGOS demonstrated here enables analogous coupling with neural network architectures tailored to partial differential equations~\cite{rudyDatadrivenDiscoveryPartial2017, martina-perezBayesianUncertaintyQuantification2021, liAutomatingDiscoveryPartial2024}, time-varying systems~\cite{liDiscoveringTimevaryingAerodynamics2019a}, and complex network dynamics~\cite{gaoLearningInterpretableDynamics2024, huLearningInterpretableNetwork2025}. Native implementations in both Python and R further broaden its accessibility across disciplines.

Like all library-based sparse regression methods, Bayesian-ARGOS remains fundamentally bounded by the expressiveness of the candidate library; accuracy cannot exceed what the provided terms permit and domain prior knowledge thus remains indispensable for constructing the hypothesis space. Additionally, extending rigorous uncertainty quantification to the frequentist screening module---thereby accounting for selection uncertainty in the full identification pipeline---remains an open challenge in post-selection inference~\cite{taylorStatisticalLearningSelective2015a}. Nevertheless, in application domains where mechanistic understanding is essential---climate modeling~\cite{gaoSparseIdentificationNonlinear2025}, neuroscience~\cite{rautArousalUniversalEmbedding2025}, and systems biology~\cite{sadriaDiscoveringGoverningEquations2025}---the framework's ability to provide calibrated uncertainty alongside discovered equations enables principled scientific inference rather than mere curve fitting. The framework's compatibility with domain-informed priors~\cite{gelmanWeaklyInformativeDefault2008, yangSymmetryInformedGoverningEquation2024} and neural latent representations further positions it as a principled bridge between physics-based and data-driven modeling paradigms. As observational datasets continue to grow in scale and complexity, the synthesis of statistical rigor, interpretability, and computational tractability demonstrated here will become increasingly central to the practice of data-driven scientific discovery.

\section{Methods}\label{sec:Methods}

Bayesian-ARGOS accelerates governing-equation discovery by integrating robust frequentist variable selection with Bayesian regression. The framework proceeds in three phases: data preprocessing via adaptive smoothing, a sequential frequentist identification module that iteratively refines the candidate library, and a Bayesian regression module for posterior sampling.

\subsection{Preprocessing and problem formulation}

To mitigate observational noise, we apply an adaptive Savitzky-Golay filter to the noisy state matrix $\tilde{\mathbf{X}} \in \mathbb{R}^{n \times m}$ to obtain smoothed states $\mathbf{X}$ and time derivative estimates $\dot{\mathbf{X}}$~\cite{eganAutomaticallyDiscoveringOrdinary2024}. The filter fits a polynomial of order $o$ over a window length $l$, where the optimal parameters $(o^*, l^*)$ are selected via grid search to minimize the cross-validated reconstruction mean squared error (MSE).

Using the smoothed states $\mathbf{X}$, we construct a candidate library $\mathbf{\Theta}(\mathbf{X}) \in \mathbb{R}^{n \times p}$ consisting of monomials up to degree $d$ and, where applicable, domain-specific nonlinear functions. The equation discovery task is then formulated as a sparse regression problem. For each state variable $x_j$ (where $j = 1, \dots, m$):
\begin{equation}
    \dot{\mathbf{x}}_j = \mathbf{\Theta}(\mathbf{X})\boldsymbol{\beta}_j + \boldsymbol{\epsilon}_j,
\end{equation}
where $\dot{\mathbf{x}}_j$ is the derivative vector of the $j$-th state, $\boldsymbol{\beta}_j$ is the sparse coefficient vector, and $\boldsymbol{\epsilon}_j$ represents the residual error. This formulation allows the use of a series of regression methods with different optimization techniques to estimate the coefficients $\boldsymbol{\beta}_j$. The underlying governing equations of the system can be determined by idenifying the optimal coefficients.

\subsection{Hybrid sequential regression framework}

Bayesian-ARGOS follows a coarse-to-fine strategy. A frequentist module first narrows the search space to a parsimonious subset of candidate terms, which is then passed to a Bayesian module for posterior inference and conditional uncertainty quantification on that subset.

\subsubsection{Two-Stage regression procedure}\label{sec:TwoStage}

The core of the frequentist module is a reusable two-stage procedure designed to eliminate false positives while correcting the shrinkage bias inherent in regularization~\cite{zouAdaptiveLassoIts2006}.

In the first stage, variable selection is performed using the adaptive LASSO regression~\cite{zouAdaptiveLassoIts2006}, which minimizes:
\begin{equation}
    \hat{\boldsymbol{\beta}}_j^{\text{adaLASSO}} = \underset{\boldsymbol{\beta}_j}{\arg\min} \left( \left\|\dot{\mathbf{x}}_j - \mathbf{\Theta}(\mathbf{X})\boldsymbol{\beta}_j\right\|_2^2 + \lambda \sum_{k=1}^p w_k |\beta_{j,k}| \right),
\end{equation}
where $\lambda > 0$ is the regularization parameter, and the penalty weights are defined as $w_k = |\hat{\beta}_{j,k}^{\text{init}}|^{-\gamma}$, with $\gamma > 0$ a tuning parameter (set to $\gamma = 1$) and $\hat{\beta}_{j,k}^{\text{init}}$ a pilot estimate obtained from a preceding regression. The efficacy of the adaptive LASSO depends critically on $\lambda$ and $\hat{\beta}_{j,k}^{\text{init}}$: the former modulates the overall penalty magnitude, while the latter determines the adaptive weights $w_k$. To optimize the selection of $\lambda$, we implement a two-phase grid search process. The initial phase involves generating a logarithmically spaced sequence of $\lambda$ values and performing 10-fold cross-validation to identify an initial optimal value $\lambda_0^*$~\cite{friedmanRegularizationPathsGeneralized2010}, as in standard implementations such as \texttt{glmnet} (R) and \texttt{Adelie} (Python). Subsequently, we conduct a refined search by constructing a denser, uniformly distributed grid of 100 points within the interval $[\lambda_{0}^* / 10, 1.1 \cdot \lambda_{0}^*]$ to obtain $\lambda^*$~\cite{eganAutomaticallyDiscoveringOrdinary2024}.

% Thresholding with OLS refiting and Information-Theoretic Selection
% and eliminate shrinkage bias inherent in regularized methods
To mitigate potential over-selection arising from the tendency of cross-validation to fit residual noise or inaccuracies in the adaptive weights $w_k$, we apply hard thresholding to $\hat{\boldsymbol{\beta}}_j^{\text{adaLASSO}}$ in the second stage~\cite{eganAutomaticallyDiscoveringOrdinary2024}. A sequence of candidate supports $\mathcal{K}_\eta = \{k : |\hat{\beta}_{j,k}^{\text{adaLASSO}}| \geq \eta\}$ is generated using a logarithmic grid of thresholds $\eta \in \{10^{-8}, \dots, 10^{1}\}$. For each $\mathcal{K}_\eta$, coefficients are re-estimated using Ordinary Least Squares (OLS) to ensure unbiasedness~\cite{belloniLeastSquaresModel2013a}. The optimal support is selected by minimizing the Bayesian Information Criterion (BIC)~\cite{schwarzEstimatingDimensionModel1978}.

\subsubsection{Sequential execution strategy}
\label{sec:sequential_execution_strategy}

The Adaptive LASSO achieves the oracle property (simultaneous consistent variable selection and optimal parameter estimation) only if the pilot estimates $\hat{\beta}_{j,k}^{\text{init}}$ reasonably approximate true predictor importance~\cite{zouAdaptiveLassoIts2006}. Standard OLS estimates provide unbiased weights but are unstable under multicollinearity, whereas Ridge estimates offer stability at the cost of bias~\cite{hastie2009elements}. To reconcile these trade-offs, we execute the two-stage procedure (Section \ref{sec:TwoStage}) twice in sequence~\cite{belloniLeastSquaresModel2013a}.

In the first execution, we utilize pilot estimates from Ridge regression ($\hat{\boldsymbol{\beta}}^{\text{ridge}}$)~\cite{hoerlRidgeRegressionBiased1970} to initialize the adaptive weights. Ridge regression effectively handles multicollinearity, ensuring that correlated potential governing terms are not prematurely excluded due to instability in the pilot estimates. This robust initial pass provides a coarse but stable identification of the system structure.

Subsequently, the design matrix undergoes strategic refinement to prevent over-regularization~\cite{eganAutomaticallyDiscoveringOrdinary2024, bienLassoHierarchicalInteractions2013}. We identify the maximum polynomial degree of any variable selected in the first execution and extend the active library to include all polynomial terms up to that degree. This safeguard ensures that true governing terms, which may have been overly penalized by conservative Ridge weights, are reconsidered in the subsequent analysis.

In the second execution, we apply the two-stage procedure to this refined library using OLS-derived pilot estimates ($\hat{\boldsymbol{\beta}}^{\text{OLS}}$), calculated on the support from the previous step. With the search space narrowed and the library safeguarded, OLS provides asymptotically unbiased weights that allow the Adaptive LASSO to converge on the true model structure~\cite{belloniLeastSquaresModel2013a}. The final sparse support identified in this stage is passed to the Bayesian module.

\subsubsection{Bayesian regression and uncertainty quantification}
\label{sec:Bayesian_regression_and_uncertainty_quantification}

The Bayesian module operates on the trimmed candidate library (reduced design matrix) identified by the frequentist block to estimate the posterior distributions of the coefficients $\boldsymbol{\beta}_j$ for each state variable.

This Bayesian approach systematically updates parameter beliefs by treating each observation as a source of information about the underlying dynamical processes. The inference process begins with prior beliefs represented by a distribution $p(\boldsymbol{\beta}_j)$ and computes posterior distributions $p(\boldsymbol{\beta}_j|\dot{\mathbf{x}}_j)$ that reflect updated beliefs after conditioning on observed data. This yields comprehensive probability distributions that capture uncertainty in both parameters and predictions, rather than point estimates alone.

According to Bayes' theorem, the posterior distribution is given by:
\begin{equation}
    p(\boldsymbol{\beta}_j|\dot{\mathbf{x}}_j) = \frac{p(\dot{\mathbf{x}}_j|\boldsymbol{\beta}_j)p(\boldsymbol{\beta}_j)}{p(\dot{\mathbf{x}}_j)},
\end{equation}
where $p(\dot{\mathbf{x}}_j|\boldsymbol{\beta}_j)$ is the likelihood function, $p(\boldsymbol{\beta}_j)$ is the prior distribution, and $p(\dot{\mathbf{x}}_j)$ is the marginal likelihood (evidence).

For each state variable $x_j$, we specify a Gaussian likelihood
\begin{equation}
    p(\dot{\mathbf{x}}_j|\boldsymbol{\beta}_j,\sigma^2) = \mathcal{N}(\dot{\mathbf{x}}_j|\mathbf{\Theta}^{(2)}(\mathbf{X})\boldsymbol{\beta}_j, \sigma^2\mathbf{I}),
\end{equation}
and assign independent priors for regression coefficients and error variance:
\begin{equation}
    p(\beta_{jk}) = \mathcal{N}(0, 2.5 s_y/s_{x_k}), \quad p(\sigma) = \text{Exp}(1/s_y).
\end{equation}
Here, $s_y$ and $s_{x_k}$ represent the sample standard deviations of the response and $k$th predictor, respectively. These weakly informative priors provide scale-appropriate regularization that automatically adjusts to the magnitudes of responses and predictors, balancing prior information with likelihood dominance to avoid over-regularization. They induce ridge-like shrinkage that enhances stability when predictors exhibit multicollinearity. The exponential prior on $\sigma$ similarly ensures robust error variance estimation through appropriate regularization.

Posterior distributions are approximated using Hamiltonian Monte Carlo (HMC) sampling implemented in \texttt{stan}~\cite{carpenterStanProbabilisticProgramming2017}. For each coefficient vector $\boldsymbol{\beta}_j$, we run four independent Markov chains with different initial values, each for 2000 iterations with the first 1000 discarded as burn-in. Convergence is assessed via the Gelman-Rubin statistic ($\hat{R} < 1.1$)~\cite{gelmanInferenceIterativeSimulation1992}.

The posterior samples enable computation of statistical quantities including posterior means and credible intervals. The final model selection employs a credibility-based criterion: variables are retained if their 90\% credible intervals exclude zero and their posterior means fall within these intervals~\cite{kruschkeRejectingAcceptingParameter2018}. This ensures that the final model reflects statistically significant effects while providing rigorous uncertainty quantification for parameter estimates.

\subsection{Experiments on benchmarking chaotic systems}

We evaluate Bayesian-ARGOS using two numerical experiments designed to characterize its sensitivity to data quantity and noise levels. For each of the seven benchmark systems (Lorenz, Thomas, Rössler, Dadras, Aizawa, Sprott, and Halvorsen), trajectories were generated using the RK45 adaptive numerical integrator~\cite{dormandFamilyEmbeddedRungeKutta1980}. To account for dynamical variability, we conducted 100 independent trials for each experimental condition, with each trial initiated from a randomly sampled initial condition.

\subsubsection{Data quantity experiments}

To characterize performance in data-limited regimes, we systematically varied the number of observations $n$ within the range $[10^2, 10^5]$ while maintaining a fixed signal-to-noise ratio (SNR) of 49~dB~\cite{bruntonDiscoveringGoverningEquations2016,eganAutomaticallyDiscoveringOrdinary2024}. Integration was performed from $t_0 = 0$ with a uniform step size $\Delta t = 0.01$, varying the final time $t_{\text{final}}$ logarithmically to achieve the desired $n$. For the Lorenz system, the step size was refined to $\Delta t = 0.001$ to capture its faster dynamics, with $t_{\text{final}}$ adjusted commensurately. This sweep allows us to identify the minimum observational threshold required for reliable model discovery across systems of varying complexity.

\subsubsection{Noise robustness experiments}

To evaluate algorithmic resilience to observational noise, we systematically varied the signal-to-noise ratio (SNR) while fixing the number of observations at $n=5000$. For each SNR level, we performed 100 trials starting from random initial conditions. Gaussian noise was added to the noise-free state variables, with the noise standard deviation for the $j$-th variable, $\sigma_{z_j}$, calibrated according to the desired SNR:
\begin{equation}
    \sigma_{z_j} = \sigma_{x_j} \cdot 10^{-\text{SNR}/20}, \quad j = 1,\ldots,m
\end{equation}
where $\sigma_{x_j}$ is the sample standard deviation of the $j$-th noise-free state variable. The noise-contaminated observations are then given by $\tilde{\mathbf{X}} = \mathbf{X} + \mathbf{Z}$, where $\mathbf{Z} \sim \mathcal{N}(0, \text{diag}(\sigma_{z_1}^2, \ldots, \sigma_{z_m}^2))$~\cite{eganAutomaticallyDiscoveringOrdinary2024}. We evaluated SNR levels ranging from 1 to 61~dB in 1~dB increments, alongside a noiseless case (SNR~=~$\infty$).

\subsubsection{Algorithms settings and evaluation metrics for performance comparison}
\label{sec: performance_comparison}

We compared Bayesian-ARGOS with ARGOS and SINDy on the benchmark systems. The ARGOS implementation employed adaptive lasso as the underlying sparse regression method, which has demonstrated effectiveness in identifying ordinary differential equations in higher dimensional systems~\cite{eganAutomaticallyDiscoveringOrdinary2024}. To ensure optimal performance, we implemented 2000 bootstrap iterations following the original specification to construct statistically valid confidence intervals~\cite{efronIntroductionBootstrap1994b}. SINDy was implemented via the \texttt{PySINDy} package through R's \texttt{reticulate} interface, using the default sequential thresholded least squares (STLSQ) optimizer with a hard threshold of 0.1 and ridge regression as the underlying regression method~\cite{bruntonDiscoveringGoverningEquations2016}.

To evaluate and compare their performance, we utilized two primary metrics: identification accuracy and computational efficiency.

Identification accuracy was quantified using the success rate, defined as the percentage of trials that correctly identified all terms in the governing equations:
\begin{equation}
    \text{Success Rate} = \frac{N_{\text{success}}}{N_{\text{total}}} \times 100\%
\end{equation}
where $N_{\text{success}}$ is the number of trials that correctly identified the exact structure of the governing equations (i.e., no false positives or false negatives) and $N_{\text{total}}$ is the total number of trials, set to 100 for all experiments. As discussed in Section~\ref{sec2}, we adopted an 80\% success rate threshold (70\% for the Aizawa system) as the benchmark for reliable identification.

Computational efficiency was assessed by recording the execution times for Bayesian-ARGOS, ARGOS, and SINDy. To accelerate the extensive suite of experiments conducted in this study, we utilized parallel computing on a high performance computing (HPC) cluster equipped with 128 CPU cores (2$\times$ AMD EPYC 7702 processors) and 256GB RAM per node. For a fair comparison of algorithmic efficiency, all multi-core processing times were converted to single-core equivalents. Representative computing times were calculated for the Lorenz and Thomas systems across logarithmically spaced observation counts $n \in [10^3, 10^5]$ with $10^{0.5}$ increments.
%  For Aizawa, we used the high-memory compute node with 128 CPU cores (2x AMD EPYC 7702), 2TB RAM

\subsection{Integration with Deep Learning for High-Dimensional Spatiotemporal Dynamics}
\label{sec:sst_sindyshred_integration}

To extend Bayesian-ARGOS to high-dimensional spatiotemporal fields observed through sparse sensors, we adopt SINDy-SHRED~\cite{gaoSparseIdentificationNonlinear2025} as a representation-learning backbone and then apply Bayesian-ARGOS \emph{post hoc} to identify governing equations in the learned latent space. SINDy-SHRED maps lagged sensor trajectories into a low-dimensional latent state via the final hidden state of a GRU module, and reconstructs the corresponding full field using a shallow decoder. Because a representation trained purely for reconstruction accuracy need not admit a parsimonious dynamical model, SINDy-SHRED introduces a differentiable SINDy-based dynamics-consistency regularizer. Concretely, the SINDy module is implemented as an RNN-form residual Euler update that advances the latent state forward by one physical time step; its prediction is matched to the subsequent GRU latent and penalized during training. This regularizer biases the learned latent coordinates toward a representation in which sparse dynamics can be expressed, while preserving the reconstruction pathway driven by the GRU module. After training, we extract the GRU latent trajectories and fit sparse latent ODEs using Bayesian-ARGOS Forecasts are then obtained by simulating the identified latent system forward from an initial latent condition and decoding the latent rollout to produce full-field predictions.

\subsubsection{Representation learning backbone}

Let $\mathbf{X}_t\in\mathbb{R}^{M}$ denote the full spatial field at time $t$, and let $\mathbf{X}^{\mathcal{S}}_{t-L+1:t}\in\mathbb{R}^{m\times L}$ collect measurements from a fixed subset of $m$ sensors over a lag window of length $L$ ending at $t$. A GRU module maps the sensor history to a latent state $\mathbf{z}^{\mathrm{GRU}}_t\in\mathbb{R}^{r}$ via its final hidden state,
\begin{equation}
    \mathbf{z}^{\mathrm{GRU}}_t
    =
    f_{\theta_G}\!\left(\mathbf{X}^{\mathcal{S}}_{t-L+1:t}\right),
\end{equation}
and a shallow decoder reconstructs the field at the terminal time of the window,
\begin{equation}
    \widehat{\mathbf{X}}_t
    =
    f_{\theta_D}\!\left(\mathbf{z}^{\mathrm{GRU}}_t\right).
\end{equation}
To regularize latent evolution, SINDy-SHRED embeds a differentiable one-step SINDy update over the interval $(t-\Delta t,\,t)$, where $\Delta t$ denotes one physical time step. For each ensemble replicate $b\in\{1,\dots,B\}$ with coefficient matrix $\Xi^{(b)}\in\mathbb{R}^{p\times r}$, SINDy posits $\dot{\mathbf{z}}=\Theta(\mathbf{z})\,\Xi^{(b)}$, where $\Theta:\mathbb{R}^{r}\rightarrow \mathbb{R}^{p}$ is a candidate function library. Given $\mathbf{z}^{\mathrm{GRU}}_{t-\Delta t}$, we compute an auxiliary rollout to time $t$ by unrolling $k$ explicit Euler mini-steps with step size $h=\Delta t/k$:
\begin{equation}
    \mathbf{z}^{\mathrm{SINDy},(b)}_{t-\Delta t}=\mathbf{z}^{\mathrm{GRU}}_{t-\Delta t},\qquad
    \mathbf{z}^{\mathrm{SINDy},(b)}_{t-\Delta t+(j+1)h}
    =
    \mathbf{z}^{\mathrm{SINDy},(b)}_{t-\Delta t+jh}
    +
    \Theta\!\left(\mathbf{z}^{\mathrm{SINDy},(b)}_{t-\Delta t+jh}\right)\,\Xi^{(b)}\,h,
\end{equation}
for $j=0,\ldots,k-1$, yielding the one-setp prediction $\mathbf{z}^{\mathrm{SINDy},(b)}_{t}$.
% In practice, training mini-batches are constructed from consecutive time indices to ensure valid $(t-\Delta t,\,t)$ pairs (e.g., we use $k=10$ Euler steps in our experiments).

Robustness is improved by evaluating an ensemble $\{\Xi^{(b)}\}_{b=1}^{B}$ in parallel. Sparsity is promoted via periodic replicate-specific hard-thresholding: at scheduled training epochs, coefficients with magnitude below a replicate-dependent threshold $\eta_b$ are masked (set to zero). The thresholds $\{\eta_b\}$ are spaced on a geometric grid to span varying sparsity levels.
% \begin{equation}
%     \eta_b = \eta_{\mathrm{base}} \cdot 10^{(0.2(b-1)-1)} + \eta_0, \qquad b=1,\dots,B.
% \end{equation}
% Robustness is improved by evaluating an ensemble $\{\Xi^{(b)}\}_{b=1}^{B}$ in parallel.
% Sparsity is promoted via replicate-specific hard thresholding: at scheduled training epochs, we apply masks
% \begin{equation}
%     \Xi^{(b)} \leftarrow \mathcal{M}^{(b)}\odot \Xi^{(b)}, \qquad
%     \mathcal{M}^{(b)}_{ij} = \mathbb{I}\!\left(|\Xi^{(b)}_{ij}| > \eta_b\right),
% \end{equation}
% where $\odot$ denotes the Hadamard product.
% The thresholds $\{\eta_b\}$ are chosen on a geometric grid across replicates to span different sparsity levels.
% In addition, AdamW weight decay is applied to stabilize optimization.

Training minimizes a composite loss comprising reconstruction error, dynamics consistency, and a latent centering term:
\begin{equation}
    \label{eq:sindyshred_loss}
    \mathcal{L}
    =
    \underbrace{\left\|\mathbf{X}_t-\widehat{\mathbf{X}}_t\right\|_2^2}_{\mathcal{L}_{\mathrm{recon}}}
    +
    \alpha\,
    \underbrace{\frac{1}{B}\sum_{b=1}^{B}\left\|\mathbf{z}^{\mathrm{GRU}}_{t}-\mathbf{z}^{\mathrm{SINDy},(b)}_{t}\right\|_2^2}_{\mathcal{L}_{\mathrm{dyn}}}
    +
    \gamma\,
    \underbrace{\left\|\boldsymbol{\mu}_z\right\|_1}_{\mathcal{L}_{\mathrm{center}}},
\end{equation}
where $\boldsymbol{\mu}_z$ denotes the empirical mean of $\mathbf{z}^{\mathrm{GRU}}$ over the current mini-batch.
Because $\mathcal{L}_{\mathrm{dyn}}$ constrains only an auxiliary one-step SINDy rollout to match the subsequent GRU latent,
the reconstruction pathway remains driven by $\mathbf{z}^{\mathrm{GRU}}_t$, reducing the risk of prematurely over-constraining the learned representation.
% Notably, sparsity is enforced through periodic hard-thresholding (and weight decay), rather than an explicit $\ell_0$/$\ell_1$ penalty on $\Xi^{(b)}$.

\subsubsection{Latent dynamics identification and forecasting}

After training, we freeze $(\theta_G,\theta_D)$ and extract a latent trajectory from the training segment. Specifically, for each time index $t$ we compute the GRU latent state
$\mathbf{z}^{\mathrm{GRU}}_t=f_{\theta_G}\!\left(\mathbf{X}^{\mathcal S}_{t-L+1:t}\right)$,
yielding $\{\mathbf{z}^{\mathrm{GRU}}_t\}$. We then apply Bayesian-ARGOS to identify a sparse latent dynamical system
$\dot{\mathbf{z}}=g(\mathbf{z})$
from this trajectory. Forecasts are produced by simulating the identified latent dynamics forward from a latent initial condition and decoding the resulting latent rollout via $\widehat{\mathbf{X}}_t=f_{\theta_D}\!\left(\widehat{\mathbf{z}}_t\right)$. As a baseline, we fit a standard SINDy model to the same latent trajectory and evaluate forecasts under the identical initialization, simulation, and decoding protocol.

\subsubsection{Experimental setup and evaluation: NOAA sea-surface temperature}

We evaluate on the NOAA SST dataset of $1{,}400$ weekly snapshots~\cite{reynoldsImprovedSituSatellite2002a}. We construct supervised samples using a sliding lag window of length $L=52$ (one year), yielding $1{,}348$ windowed samples. For each window start index $i$, the input is the sensor trajectory $\mathbf{X}^{\mathcal S}_{i:i+L-1}\in\mathbb{R}^{L\times m}$ and the reconstruction target is the full field at the terminal week $\mathbf{X}_{i+L-1}$. We use the first 1000 samples for training, the next 30 for validation, and the remaining 318 for testing. Each trial uses $m=250$ sensors selected uniformly at random and re-sampled across trials.

The model comprises a two-layer GRU with latent dimension $r=3$ and a two-layer ReLU decoder with 350 and 400 hidden units. We regularize latent dynamics using an ensemble of $B=10$ coefficient replicates and a degree-3 polynomial library. The SINDy module is implemented as an Euler residual update unrolled for $k=10$ mini-steps with $h=\texttt{dt}=1/520$, yielding an effective step size $\Delta t=kh=1/52$ (one week). Training uses AdamW (learning rate $10^{-3}$, weight decay $10^{-2}$) with batch size 128, and applies periodic coefficient hard-thresholding every 100 epochs, for up to 600 epochs with validation monitored every 100 epochs. We repeat the full pipeline for 150 trials, utilizing an NVIDIA A100 80GB GPU on a GPU-accelerated HPC cluster.

To decouple representation errors from downstream identification errors, we retain only runs whose test-set reconstruction error is below $0.05$, computed as a relative $\ell_2$ error over aggregated test spatiotemporal points:
\begin{equation}
    \label{eq:rel_l2_recon}
    \mathrm{rRMSE}
    \;=\;
    \frac{\left\|\widehat{\mathbf{y}}-\mathbf{y}\right\|_2}{\left\|\mathbf{y}\right\|_2},
\end{equation}
where $\mathbf{y}$ and $\widehat{\mathbf{y}}$ are the vectorized ground-truth and reconstructed fields over the test set.
% where $\mathbf{y}=\mathrm{vec}(\mathbf{X})$ and $\widehat{\mathbf{y}}=\mathrm{vec}(\widehat{\mathbf{X}})$ stack the ground-truth and reconstructed SST values over the full test set.
For retained models, we evaluate identification and forecasting accuracy by simulating the identified latent dynamics from a latent initial condition and decoding the latent rollout to the full field; we report latent-space MSE and full-field RMSE. As a baseline, we fit SINDy using \texttt{PySINDy} with a first-order polynomial library, finite-difference differentiation, and the STLSQ optimizer (threshold $0.8$, ridge parameter $\alpha=0.05$), and evaluate forecasts under the same initialization, simulation, and decoding protocol.

\subsection{Data availability}
\label{subsect:data}
\sloppy All simulated data generated for this research can be generated with the code available at \url{https://github.com/yuzhengv/Bayesian-ARGOS}. The Sea Surface Temperature (SST) dataset can be downloaded from \url{https://drive.google.com/file/d/1IrKFsYEcUL8xxZ0PUSLC3VrpTvVneDhj/view}.

\subsection{Code availability}
\label{subsect:code_avail}
The method has been implemented in both R and Python. The R version is available at \url{https://github.com/yuzhengv/Bayesian-ARGOS} and was used for performance evaluation. The Python version is available at \url{https://github.com/yuzhengv/pyargos_with_sindy_shred} and includes integration with SINDy-SHRED and diagnostic analyses.

\backmatter

\bmhead{Acknowledgements}

Y.Z. acknowledges support from the Engineering and Physical Sciences Research Council (EPSRC) [EP/W524426/1] (studentship project reference 2744980).
This work has made use of Durham University's Hamilton HPC and NCC clusters. NCC has been purchased through Durham University's strategic investment funds, and is installed and maintained by the Department of Computer Science. We gratefully acknowledge Kevin Egan for his contributions to the preliminary analysis that informed this work.

\bmhead{Author contributions}

Y.Z. conducted the research, designed the methodology, implemented the code, performed the experiments and validation tests, and wrote the manuscript.
W.L. contributed to the methodology design and participated in discussions of the experimental design and validation results.
R.C. supervised the project and reviewed and revised the manuscript.

\bmhead{Supplementary information}

The online version of this article includes supplementary information.

\clearpage

% \begin{appendices}

%     \section{Section title of first appendix}\label{secA1}

%     %%=============================================%%
%     %% For submissions to Nature Portfolio Journals %%
%     %% please use the heading ``Extended Data''.   %%
%     %%=============================================%%

%     %%=============================================================%%
%     %% Sample for another appendix section			       %%
%     %%=============================================================%%

%     %% \section{Example of another appendix section}\label{secA2}%
%     %% Appendices may be used for helpful, supporting or essential material that would otherwise 
%     %% clutter, break up or be distracting to the text. Appendices can consist of sections, figures, 
%     %% tables and equations etc.

% \end{appendices}


\bibliography{references2,  references1}

\clearpage
\newcommand{\beginsupplement}{%
    \section*{Supplementary Information}\label{Supplementary_Information}
    \setcounter{section}{0}
    \renewcommand{\thesection}{S\arabic{section}}
    \renewcommand{\theHsection}{S\arabic{section}}
    \newcounter{SIsec}
    \renewcommand{\theSIsec}{S\arabic{SIsec}}
    \setcounter{table}{0}
    \renewcommand{\thetable}{S\arabic{table}}
    \renewcommand{\theHtable}{S\arabic{table}}
    \newcounter{SItab}
    \renewcommand{\theSItab}{S\arabic{SItab}}
    \setcounter{figure}{0}
    \renewcommand{\thefigure}{S\arabic{figure}}
    \renewcommand{\theHfigure}{S\arabic{figure}}
    \newcounter{SIfig}
    \renewcommand{\theSIfig}{S\arabic{SIfig}}
    \setcounter{equation}{0}
    \renewcommand{\theequation}{S\arabic{equation}}
    \renewcommand{\theHequation}{S\arabic{equation}}
    \newcounter{SIeq}
    \renewcommand{\theSIeq}{S\arabic{SIeq}}
}

\begin{bibunit}[sn-mathphys-num]
    \beginsupplement

    \section{Success-rate benchmarking for governing-equation identification on Sprott and Halvorsen systems}

    \begin{figure}[!htbp]
        \centering
        \includegraphics[width=\textwidth]{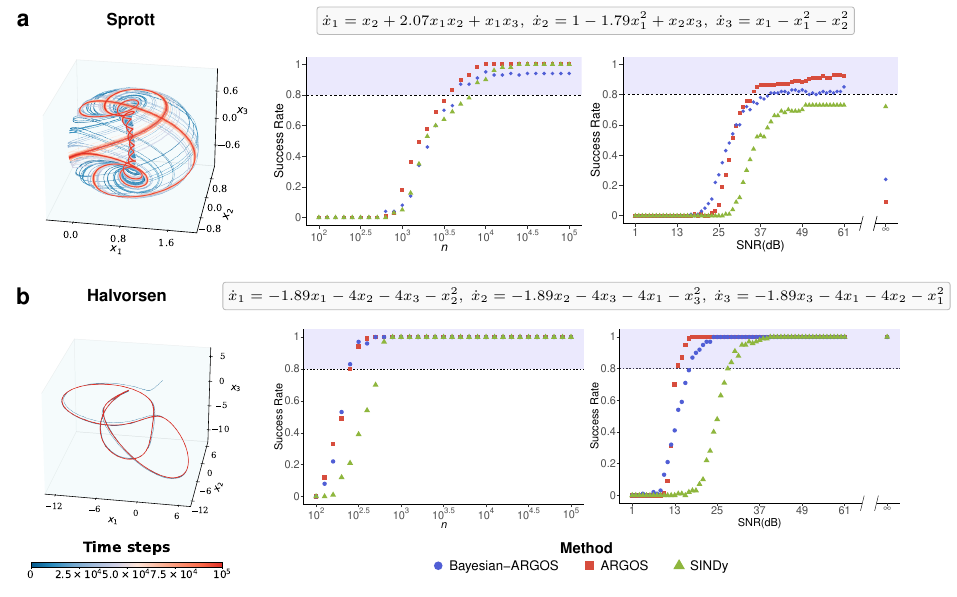}
        \caption{\textbf{Success-rate benchmarking for governing-equation identification on two chaotic systems.} \textbf{a}--\textbf{b} correspond to the Sprott and Halvorsen systems, respectively. For each system, the left subpanel shows a representative attractor; the middle subpanel shows the success rate (fraction of 100 trials recovering all ground-truth terms) versus the number of observations $n$ at SNR$=49$~dB; and the right subpanel shows the success rate versus SNR (dB) at $n=5000$. Success rate for Bayesian-ARGOS, ARGOS, and SINDy are shown in blue, red, and green, respectively. The dashed line marks the 80\% success-rate threshold.}
        \label{fig:Stacked_Identification_Results_3}
    \end{figure}

    \section{Experimental setup and evaluation on the seven chaotic systems}
    \label{supp_sect:benchmark_experimental_setup}

    \subsection*{Lorenz system}
    \label{subsupp_sect:Lorenz}
\begin{figure*}[!htbp]
        \centering
        \includegraphics[width=1\textwidth]{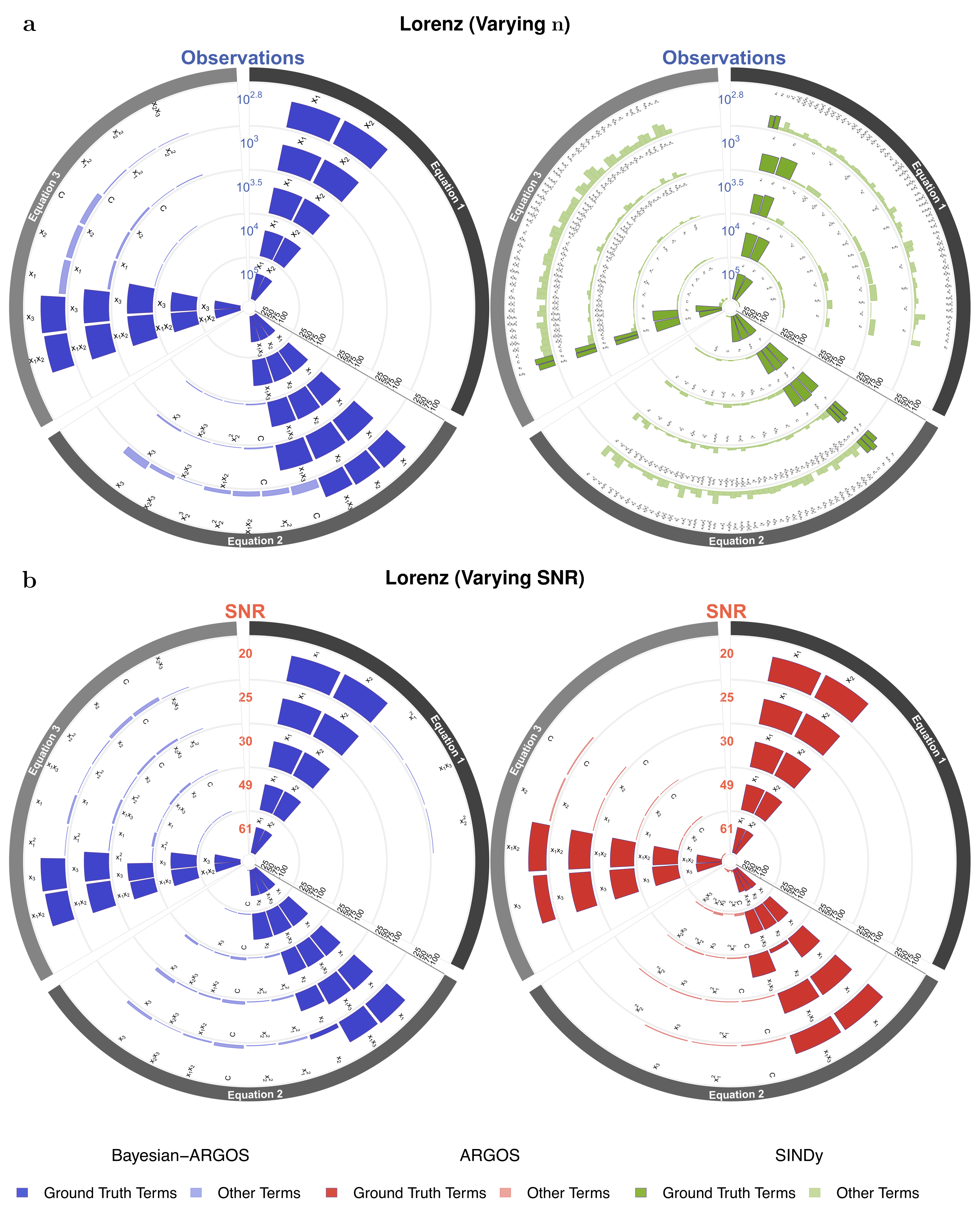}
        \caption{\textbf{Term-selection frequency on the Lorenz system.} Radial bars show how often each library term is selected across trials for each equation; dark colors denote ground-truth terms and light colors denote other terms. \textbf{a} varying the number of observations; \textbf{b} varying SNR. Bayesian-ARGOS (blue), ARGOS (red), and SINDy (green).}
        \label{fig:si_distribution_lorenz}
    \end{figure*}

    We considered the Lorenz system, a canonical low-dimensional chaotic model originally proposed to describe atmospheric convection.
    Its dynamics are governed by
    \begin{equation}
        \label{eq:lorenz_system}
        \begin{aligned}
            \dot{x}_1 & = \sigma (x_2 - x_1),    \\
            \dot{x}_2 & = x_1(\rho - x_3) - x_2, \\
            \dot{x}_3 & = x_1 x_2 - \zeta x_3,
        \end{aligned}
    \end{equation}
    We adopted the standard parameter values $\sigma=10$, $\rho=28$, and $\zeta=8/3$~\cite{bruntonDiscoveringGoverningEquations2016,eganAutomaticallyDiscoveringOrdinary2024}.
    We include the Lorenz system as a canonical benchmark because it exhibits chaos with comparatively low-order polynomial nonlinearities (primarily bilinear terms), providing a well-established testbed for evaluating sparse identification under measurement noise.
    We performed 100 trials with initial conditions sampled independently from uniform distributions: $x_1(0)\sim\mathcal{U}[-15,15]$, $x_2(0)\sim\mathcal{U}[-15,15]$, and $x_3(0)\sim\mathcal{U}[10,40]$.
    Figure~\ref{fig:si_distribution_lorenz} summarizes the selection frequency of each library term across trials as the number of observations and SNR vary. SINDy exhibits weak shrinkage, repeatedly activating non-ground-truth terms (particularly at lower data quality), whereas the stronger shrinkage of ARGOS can induce false negatives in a term-specific manner---most notably, it frequently drops the true linear damping term $-x_2$ in $\dot{x}_2$ at low-to-moderate SNR. By contrast, Bayesian-ARGOS largely preserves the ground-truth support while keeping spurious selections rare.
    % and its recovery improves more smoothly with increasing SNR rather than exhibiting the more abrupt, threshold-like omissions seen in ARGOS.

    \subsection*{Thomas system}
    \label{subsupp_sect:Thomas}
\begin{figure*}[!htbp]
        \centering
        \includegraphics[width=1\textwidth]{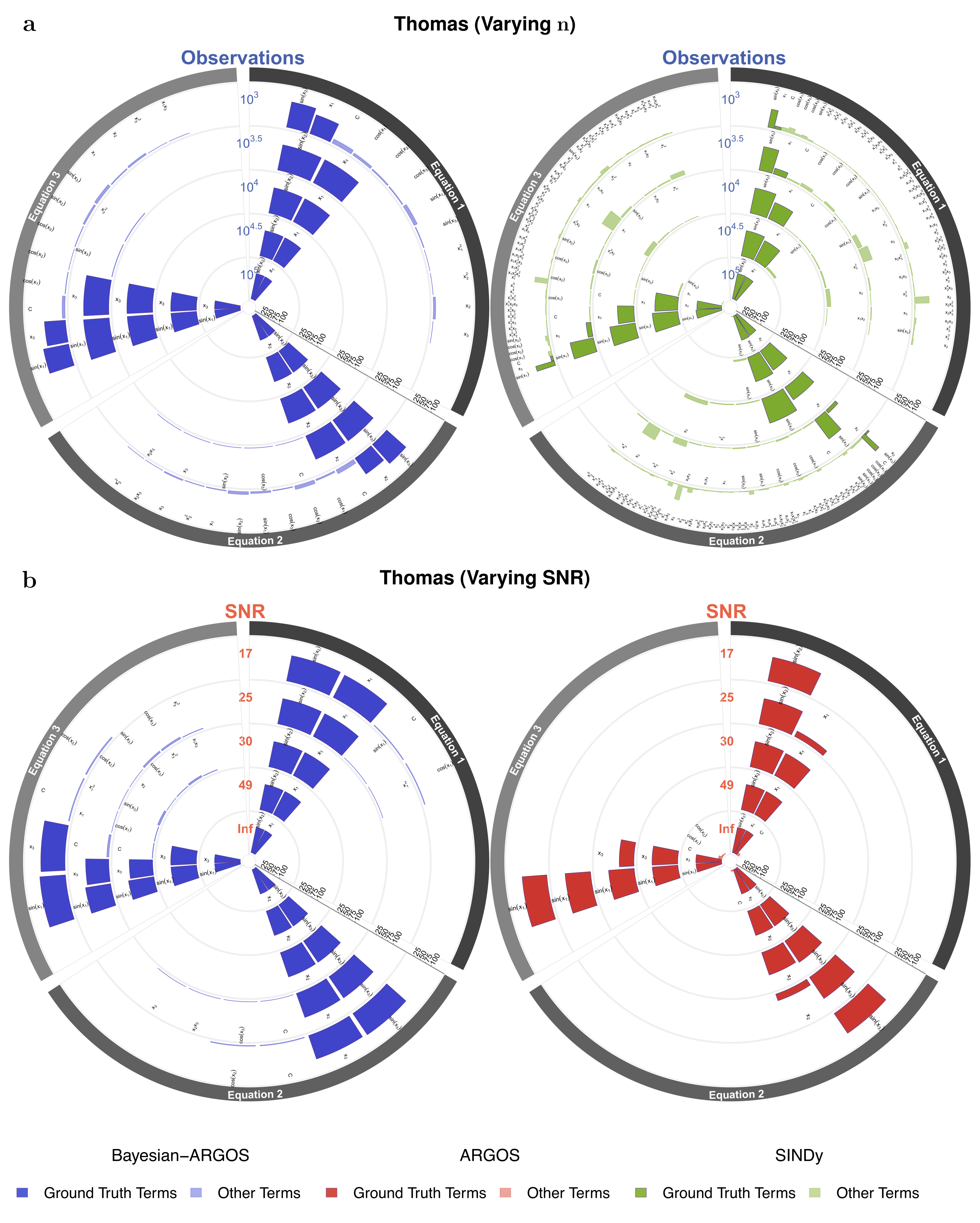}
        \caption{\textbf{Term-selection frequency on the Thomas system.} Radial bars show how often each library term is selected across trials for each equation; dark colors denote ground-truth terms and light colors denote other terms. \textbf{a} varying the number of observations; \textbf{b} varying SNR. Bayesian-ARGOS (blue), ARGOS (red), and SINDy (green).}
        \label{fig:si_distribution_thomas}
    \end{figure*}

    We examined the Thomas system, a three-dimensional chaotic system with cyclic trigonometric coupling.
    Its dynamics are governed by
    \begin{equation}
        \label{eq:thomas_system}
        \begin{aligned}
            \dot{x}_1 & = \sin(x_2) - a x_1, \\
            \dot{x}_2 & = \sin(x_3) - a x_2, \\
            \dot{x}_3 & = \sin(x_1) - a x_3,
        \end{aligned}
    \end{equation}
    We used $a = 0.208186$~\cite{sprottLabyrinthChaos2007a}. We include the Thomas system as a benchmark because its non-polynomial trigonometric nonlinearities typically require richer (and more correlated) function libraries than polynomial bases, making sparse recovery more challenging.
    To generate diverse trajectories, we performed 100 trials with initial conditions sampled independently from uniform distributions: $x_1(0)\sim\mathcal{U}[-1,1]$, $x_2(0)\sim\mathcal{U}[-1,1]$, and $x_3(0)\sim\mathcal{U}[-1,1]$.
    Figure~\ref{fig:si_distribution_thomas} summarizes the selection frequency of each library term across trials as the number of observations and SNR vary. Across regimes, Bayesian-ARGOS concentrates selection on the ground-truth cyclic structure---the driving trigonometric terms $\sin(x_2)$, $\sin(x_3)$, and $\sin(x_1)$ and the linear damping terms $x_1$, $x_2$, and $x_3$ in their respective equations---while keeping non-ground-truth terms rare. By contrast, SINDy often recovers the driving trigonometric terms but more frequently misses or mis-selects the linear damping terms when using the enriched trigonometric library, whose highly correlated candidates exacerbate multicollinearity, especially when the number of observations is small to moderate. Similar to other systems, ARGOS's stronger shrinkage can yield term-specific false negatives at lower data quality, most notably by omitting the weaker linear damping terms at low-to-moderate SNR (17--30~dB).

    \subsection*{R\"ossler system}
    \label{subsupp_sect:rossler}
\begin{figure*}[!htbp]
        \centering
        \includegraphics[width=1\textwidth]{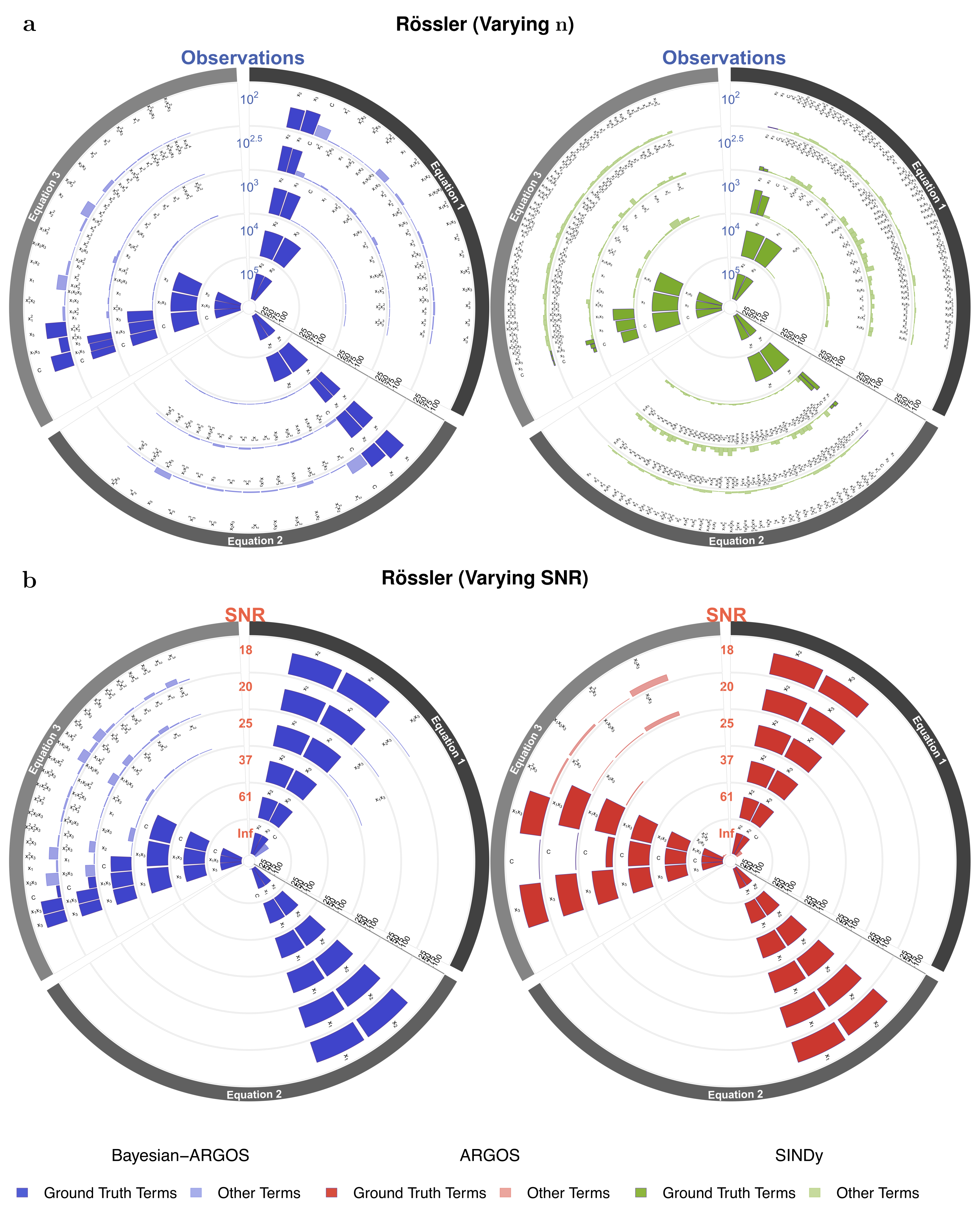}
        \caption{\textbf{Term-selection frequency on the R\"ossler system.} Radial bars show how often each library term is selected across trials for each equation; dark colors denote ground-truth terms and light colors denote other terms. \textbf{a} varying the number of observations; \textbf{b} varying SNR. Bayesian-ARGOS (blue), ARGOS (red), and SINDy (green).}
        \label{fig:si_distribution_rossler}
    \end{figure*}

    We examined the R\"ossler system, a three-dimensional chaotic system with a relatively compact low-order structure.
    Its dynamics are governed by
    \begin{equation}
        \label{eq:rossler_system}
        \begin{aligned}
            \dot{x}_1 & = -x_2 - x_3,       \\
            \dot{x}_2 & = x_1 + a x_2,      \\
            \dot{x}_3 & = b + x_3(x_1 - c),
        \end{aligned}
    \end{equation}
    We used $a = 0.2$, $b = 0.2$, and $c = 5.7$~\cite{tranExactRecoveryChaotic2017}. We include the R\"ossler system as a benchmark because it exhibits sparse parameterization with few active terms while containing multiplicative coupling (e.g., $x_1 x_3$), thereby enabling evaluation of identification performance on low-order nonlinear interactions.
    To generate diverse trajectories, we performed 100 trials with initial conditions sampled independently from uniform distributions: $x_1(0)\sim\mathcal{U}[-10,10]$, $x_2(0)\sim\mathcal{U}[-10,10]$, and $x_3(0)\sim\mathcal{U}[0,20]$. Figure~\ref{fig:si_distribution_rossler} summarizes the selection frequency of each library term across trials as the number of observations and SNR vary. Across regimes, Bayesian-ARGOS concentrates selection on the ground-truth structure---the linear terms in $\dot{x}_1$ and $\dot{x}_2$, and in $\dot{x}_3$ the constant forcing $b$, the linear $x_3$ term, and the bilinear interaction $x_1x_3$---relatively earlier than the competing methods. At $n = 10^{2}$, SINDy struggles to distinguish the true structure from spurious terms, and from $10^{2.5}$ to $10^{3.5}$ it more frequently activates additional polynomial terms, reflecting the limitations of deterministic thresholding and weaker regularization. Conversely, ARGOS's stronger shrinkage can occasionally induce false negatives for individual terms at lower data quality, most notably by omitting the constant forcing term in $\dot{x}_3$, though recovery becomes more reliable as SNR improves. This behavior is consistent with observations across other systems.

    \subsection*{Dadras system}
    \label{subsupp_sect:dadras}
    We examined the Dadras system~\cite{dadrasNovelThreedimensionalAutonomous2009}, a three-dimensional chaotic system with extensive bilinear coupling.
    Its dynamics are governed by
    \begin{equation}
        \label{eq:dadras_system}
        \begin{aligned}
            \dot{x}_1 & = x_2 - 3 x_1 + 2.7 x_2 x_3, \\
            \dot{x}_2 & = 1.7 x_2 - x_1 x_3 + x_3,   \\
            \dot{x}_3 & = 2 x_1 x_2 - 9 x_3,
        \end{aligned}
    \end{equation}
    We include the Dadras system as a benchmark because it features multiple bilinear interaction terms distributed across different variable pairs ($x_2 x_3$, $x_1 x_3$, and $x_1 x_2$), combined with linear damping in each equation. This structure tests whether the identification method can correctly distinguish and select among several competing quadratic interactions while maintaining sparsity in the recovered model.
    To generate diverse trajectories, we performed 100 trials with initial conditions sampled independently from uniform distributions: $x_1(0)\sim\mathcal{U}[-4,4]$, $x_2(0)\sim\mathcal{U}[-4,4]$, and $x_3(0)\sim\mathcal{U}[-4,4]$.

\begin{figure*}[!htbp]
        \centering
        \includegraphics[width=1\textwidth]{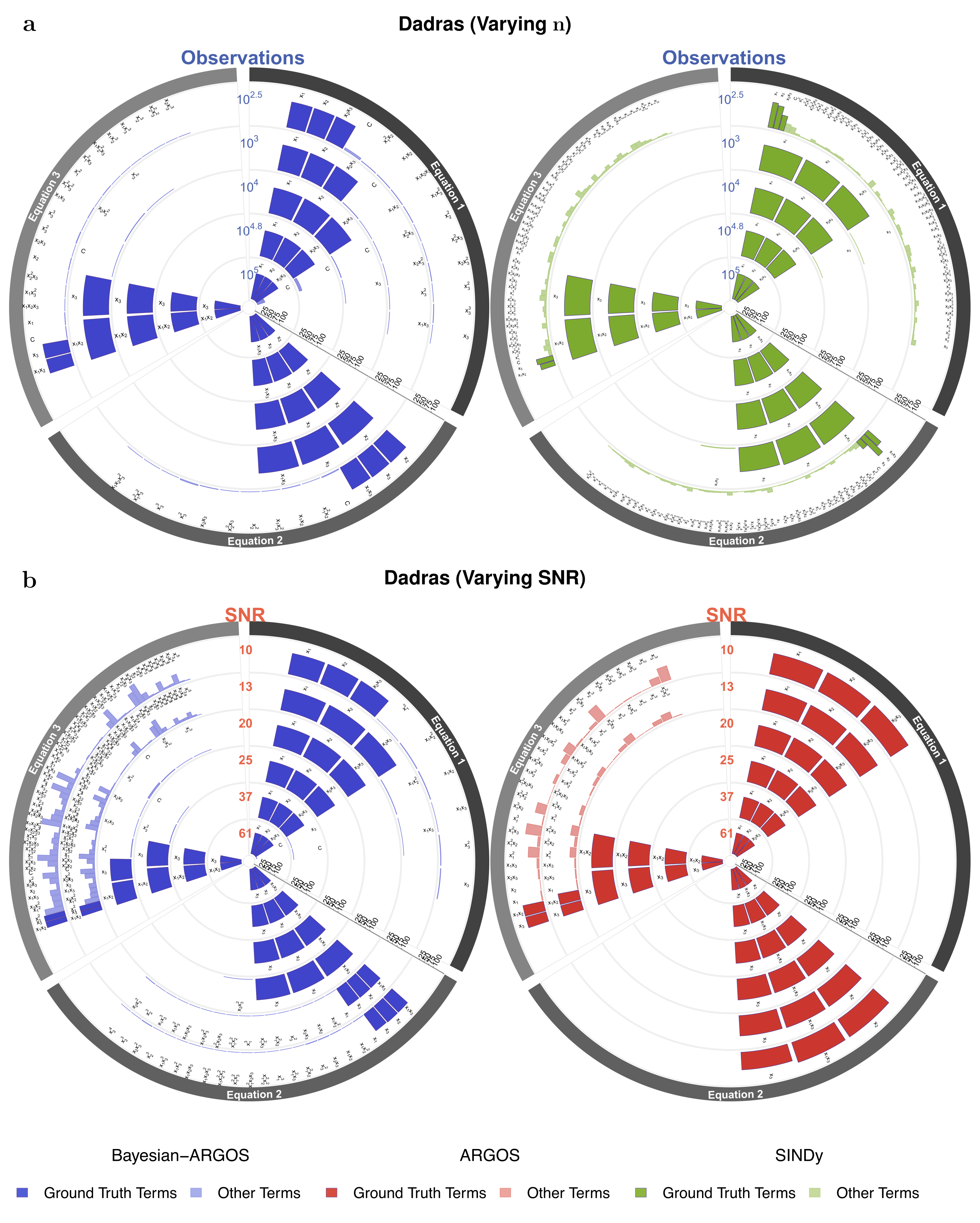}
        \caption{\textbf{Term-selection frequency on the Dadras system.} Radial bars show how often each library term is selected across trials for each equation; dark colors denote ground-truth terms and light colors denote other terms. \textbf{a} varying the number of observations; \textbf{b} varying SNR. Bayesian-ARGOS (blue), ARGOS (red), and SINDy (green).}
        \label{fig:si_distribution_dadras}
    \end{figure*}

    Figure~\ref{fig:si_distribution_dadras} summarizes the frequency with which each candidate term is selected across trials as $n$ and SNR vary. Under moderate to high observations and SNR conditions, all three algorithms converge on the Dadras ground-truth structure: linear contributions and the bilinear interaction $x_2x_3$ in $\dot{x}_1$, the mixed bilinear term $x_1x_3$ with linear terms in $\dot{x}_2$, and the $x_1x_2$ interaction with linear damping in $\dot{x}_3$. This demonstrates the capacity of these methods to discriminate among multiple competing cross-products in this strongly coupled system. As shown in Fig.~\ref{fig:si_distribution_dadras}\textbf{a}, SINDy tends to activate additional polynomial terms more frequently at $10^{2.5}$ observations. In Fig.~\ref{fig:si_distribution_dadras}\textbf{b}, both Bayesian-ARGOS and ARGOS progressively converge onto the ground-truth terms as SNR increases. However, at low SNR (10--20), Bayesian-ARGOS exhibits a modestly broader selection distribution with more spurious activations compared to ARGOS, consistent with the small success-rate gap reported in the main paper. This behavior reflects a deliberate design choice: Bayesian-ARGOS employs balanced shrinkage that is strong enough to suppress most false positives while avoiding overly aggressive sparsification that could induce term-specific false negatives under severe noise. The occasional selection of an intercept term in $\dot{x}_1$ at $n=10^5$ is consistent with the performance degradation caused by influential outliers discussed in the main paper.

    \subsection*{Aizawa system}
    \label{subsupp_sect:aizawa}
    We examined the Aizawa system~\cite{dascoli2024odeformer}, a three-dimensional chaotic system with higher-order polynomial nonlinearities.
    Its dynamics are governed by
    \begin{equation}
        \label{eq:aizawa_system}
        \begin{aligned}
            \dot{x}_1 & = -3.5 x_2 + x_1 (-0.7 + x_3),                                                                  \\
            \dot{x}_2 & = 3.5 x_1 + x_2 (-0.7 + x_3),                                                                   \\
            \dot{x}_3 & = 0.95 x_3 + 0.65 + 0.1 x_1^{3} x_3 - \tfrac{1}{3} x_3^{3} - (x_1^{2} + x_2^{2})(0.25 x_3 + 1),
        \end{aligned}
    \end{equation}
    We include the Aizawa system as a benchmark because it contains higher-order terms (e.g., $x_1^{3} x_3$ and $x_3^{3}$) and mixed products, probing identification performance when the true dynamics are not limited to low-degree polynomials.
    To generate diverse trajectories, we performed 100 trials with initial conditions sampled independently from uniform distributions: $x_1(0)\sim\mathcal{U}[-2,2]$, $x_2(0)\sim\mathcal{U}[-2,2]$, and $x_3(0)\sim\mathcal{U}[-1,2]$.

\begin{figure*}[!htbp]
        \centering
        \includegraphics[width=1\textwidth]{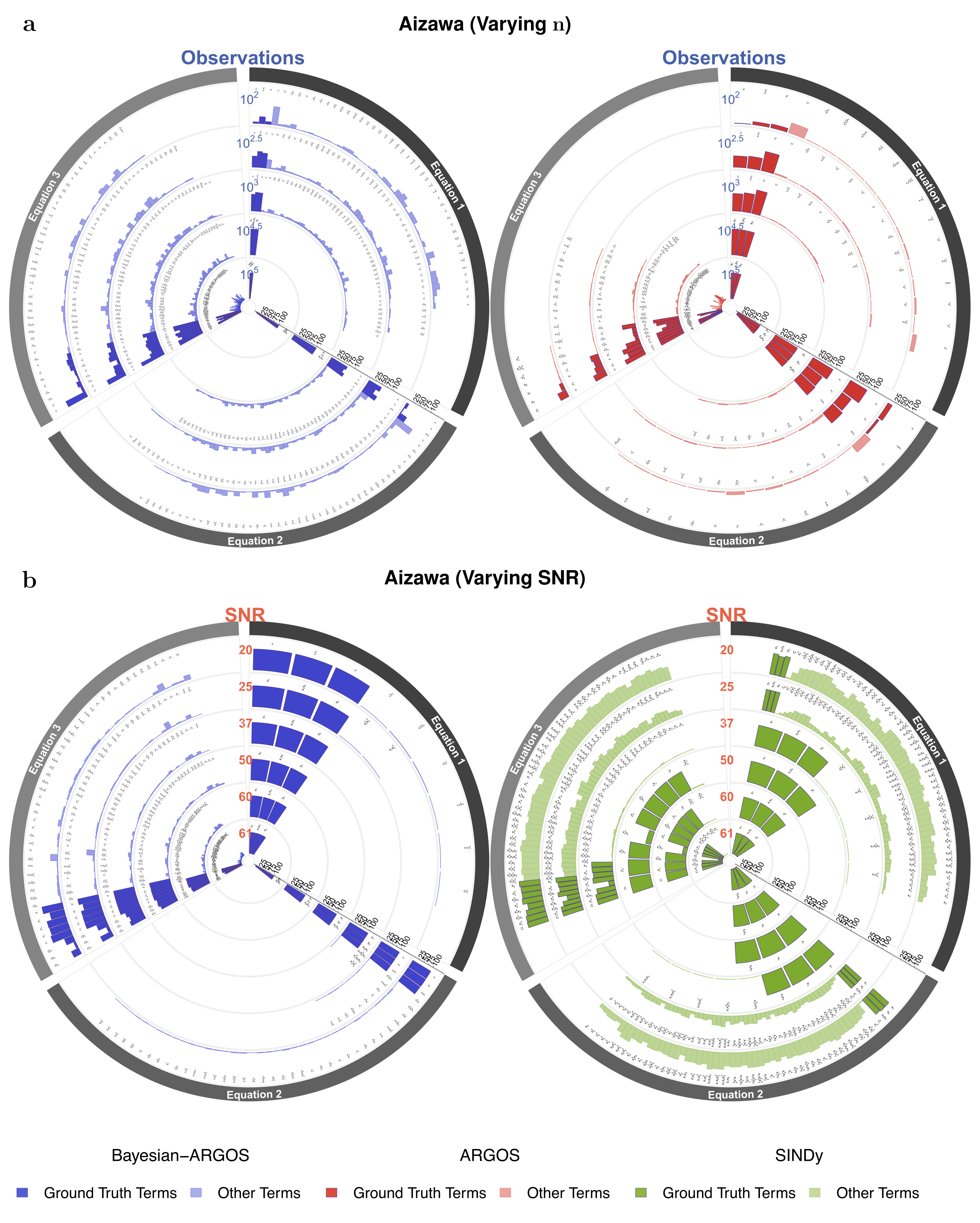}
        \caption{\textbf{Term-selection frequency on the Aizawa system.} Radial bars show how often each library term is selected across trials for each equation; dark colors denote ground-truth terms and light colors denote other terms. \textbf{a} varying the number of observations; \textbf{b} varying SNR. Bayesian-ARGOS (blue), ARGOS (red), and SINDy (green).}
        \label{fig:si_distribution_aizawa}
    \end{figure*}

    Figure~\ref{fig:si_distribution_aizawa} summarizes the frequency with which each candidate term is selected across trials as $n$ and SNR vary. For the relatively simple equations $\dot{x}_1$ and $\dot{x}_2$, both Bayesian-ARGOS and ARGOS concentrate selection on the ground-truth structure under moderate to high observations, while largely avoiding spurious activations. In contrast, the $\dot{x}_3$ equation presents a substantially more challenging identification problem because it combines constant forcing, a cubic self-term ($x_3^3$), and several correlated higher-order products (e.g., $x_1^3x_3$, $x_1^2$, $x_2^2$, $x_1^2x_3$, and $x_2^2x_3$). This structural complexity is reflected in broader selection distributions across all observation counts in Fig.~\ref{fig:si_distribution_aizawa}\textbf{a}. As shown in Fig.~\ref{fig:si_distribution_aizawa}\textbf{b}, the methods exhibit contrasting behavior as SNR varies: Bayesian-ARGOS progressively concentrates on the full ground-truth support as SNR increases, though it cannot avoid spurious activations caused by multicollinearity (as discussed in the paper). SINDy, however, exhibits overselection across the enriched polynomial library at lower SNR (20 and 30~dB) but then typically omits the $x_1^3x_3$ term in $\dot{x}_3$ at higher SNR (40 and 50~dB), demonstrating inconsistent term selection behavior under varying noise conditions.

    \subsection*{Sprott system}
    \label{subsupp_sect:sprott}
    We examined the Sprott system~\cite{sprottDynamicalSystemStrange2014}, a three-dimensional chaotic system characterized by the coexistence of constant forcing terms and multiple quadratic nonlinear interactions.
    Its dynamics are governed by
    \begin{equation}
        \label{eq:sprott_system}
        \begin{aligned}
            \dot{x}_1 & = x_2 + 2.07 x_1 x_2 + x_1 x_3, \\
            \dot{x}_2 & = 1 - 1.79 x_1^{2} + x_2 x_3,   \\
            \dot{x}_3 & = x_1 - x_1^{2} - x_2^{2},
        \end{aligned}
    \end{equation}
    The Sprott system presents a distinctive identification challenge by combining nonzero constant terms (intercepts representing external forcing) with diverse quadratic interactions that include both bilinear cross-products ($x_1 x_2$, $x_1 x_3$, $x_2 x_3$) and self-interaction terms ($x_1^{2}$, $x_2^{2}$). This structure probes the method's ability to simultaneously identify forcing components and distinguish among multiple competing nonlinear terms under observational noise, testing whether the regression framework can maintain correct term selection across different functional forms within a single system.
    To generate diverse trajectories, we performed 100 trials with initial conditions sampled independently from uniform distributions: $x_1(0)\sim\mathcal{U}[-1,1]$, $x_2(0)\sim\mathcal{U}[-1,1]$, and $x_3(0)\sim\mathcal{U}[-1,1]$.

\begin{figure*}[!htbp]
        \centering
        \includegraphics[width=1\textwidth]{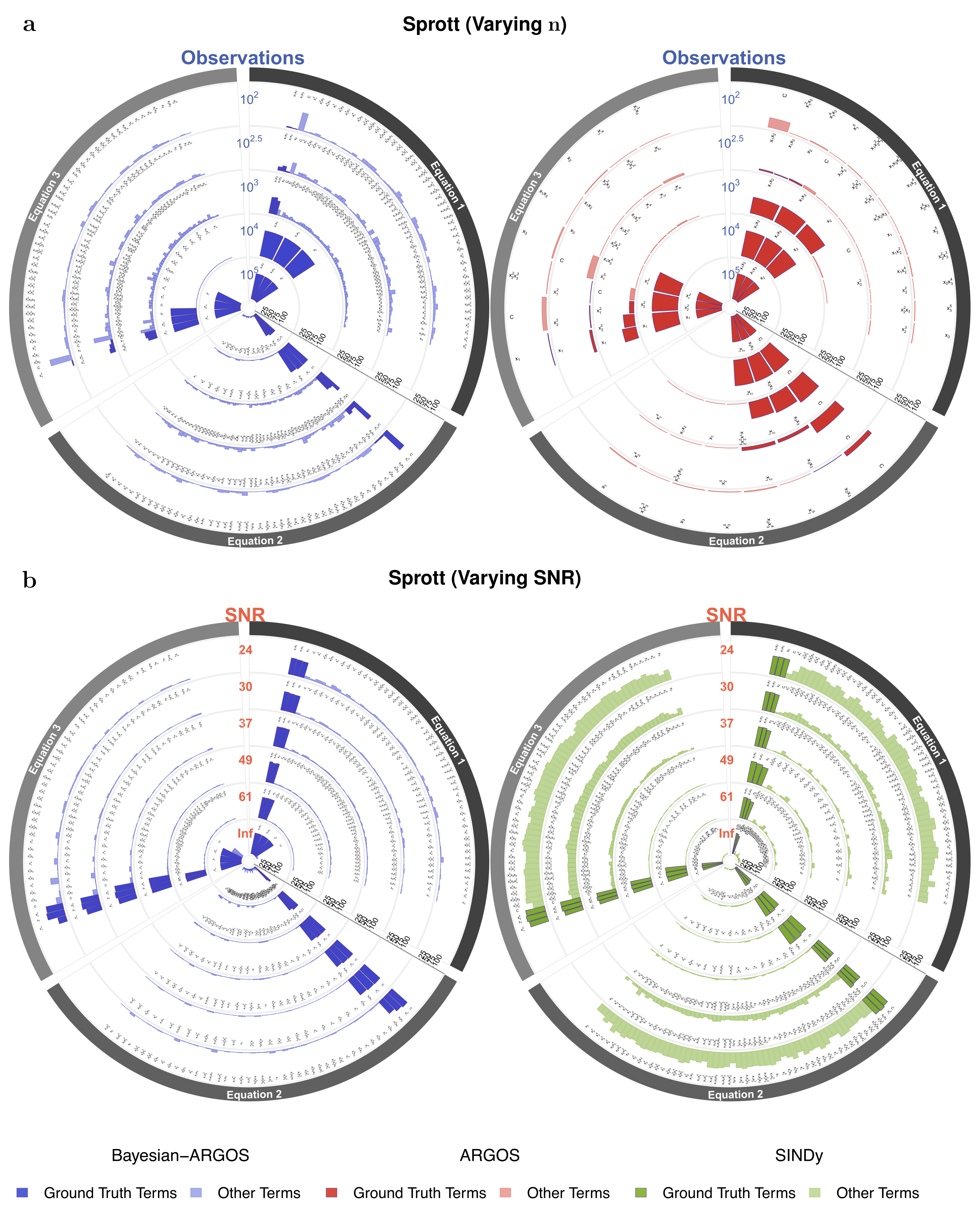}
        \caption{\textbf{Term-selection frequency on the Sprott system.} Radial bars show how often each library term is selected across trials for each equation; dark colors denote ground-truth terms and light colors denote other terms. \textbf{a} varying the number of observations; \textbf{b} varying SNR. Bayesian-ARGOS (blue), ARGOS (red), and SINDy (green).}
        \label{fig:si_distribution_sprott}
    \end{figure*}

    Figure~\ref{fig:si_distribution_sprott} summarizes the frequency with which each candidate term is selected across trials as $n$ and SNR vary. As the number of observations increases (Fig.~\ref{fig:si_distribution_sprott}\textbf{a}), ARGOS progressively concentrates its selection on the ground-truth support across all three equations. In contrast, Bayesian-ARGOS's balanced shrinkage yields a generally broader selection distribution when data are limited. Notably, the constant forcing term in $\dot{x}_3$ is more frequently activated than the ground-truth term from $n=10^{2}$ to $n=10^{3}$, and this behavior also causes spurious terms to appear in the $\dot{x}_2$ equation at $n=10^{5}$. As SNR increases (Fig.~\ref{fig:si_distribution_sprott}\textbf{b}), Bayesian-ARGOS sharpens its selection distribution toward the true support, but frequently includes spurious terms when SNR is infinite, which are partly caused by violation of the homoscedasticity assumption as discussed in the paper. In comparison, SINDy exhibits more frequent activation of non-ground-truth polynomial terms at low-to-moderate SNR and also can not fully avoid spurious terms when SNR is infinite, indicating that the Sprott system poses a common challenge for sparse identification methods under various noise conditions.

    \subsection*{Halvorsen system}
    \label{subsupp_sect:halvorsen}
    We examined the Halvorsen system~\cite{sprottChaosTimeSeriesAnalysis2003}, a three-dimensional chaotic system characterized by fully coupled linear interactions and cyclic quadratic self-terms.
    Its dynamics are governed by
    \begin{equation}
        \label{eq:halvorsen_system}
        \begin{aligned}
            \dot{x}_1 & = -1.89 x_1 - 4 x_2 - 4 x_3 - x_2^{2}, \\
            \dot{x}_2 & = -1.89 x_2 - 4 x_3 - 4 x_1 - x_3^{2}, \\
            \dot{x}_3 & = -1.89 x_3 - 4 x_1 - 4 x_2 - x_1^{2},
        \end{aligned}
    \end{equation}
    The Halvorsen system presents a distinctive identification challenge through its combination of dense linear coupling (each equation depends linearly on all three state variables) and cyclic quadratic self-terms ($x_2^{2}$ in the first equation, $x_3^{2}$ in the second, and $x_1^{2}$ in the third). This structure probes the method's ability to correctly differentiate between linear cross-coupling and self-interaction nonlinearities, testing whether the sparse regression framework can maintain accurate term selection when multiple linear terms compete for significance alongside strategically positioned quadratic terms that do not involve bilinear cross-products.
    To generate diverse trajectories, we performed 100 trials with initial conditions sampled independently from uniform distributions: $x_1(0)\sim\mathcal{U}[-4,4]$, $x_2(0)\sim\mathcal{U}[-4,4]$, and $x_3(0)\sim\mathcal{U}[-4,4]$.

\begin{figure*}[!htbp]
        \centering
        \includegraphics[width=1\textwidth]{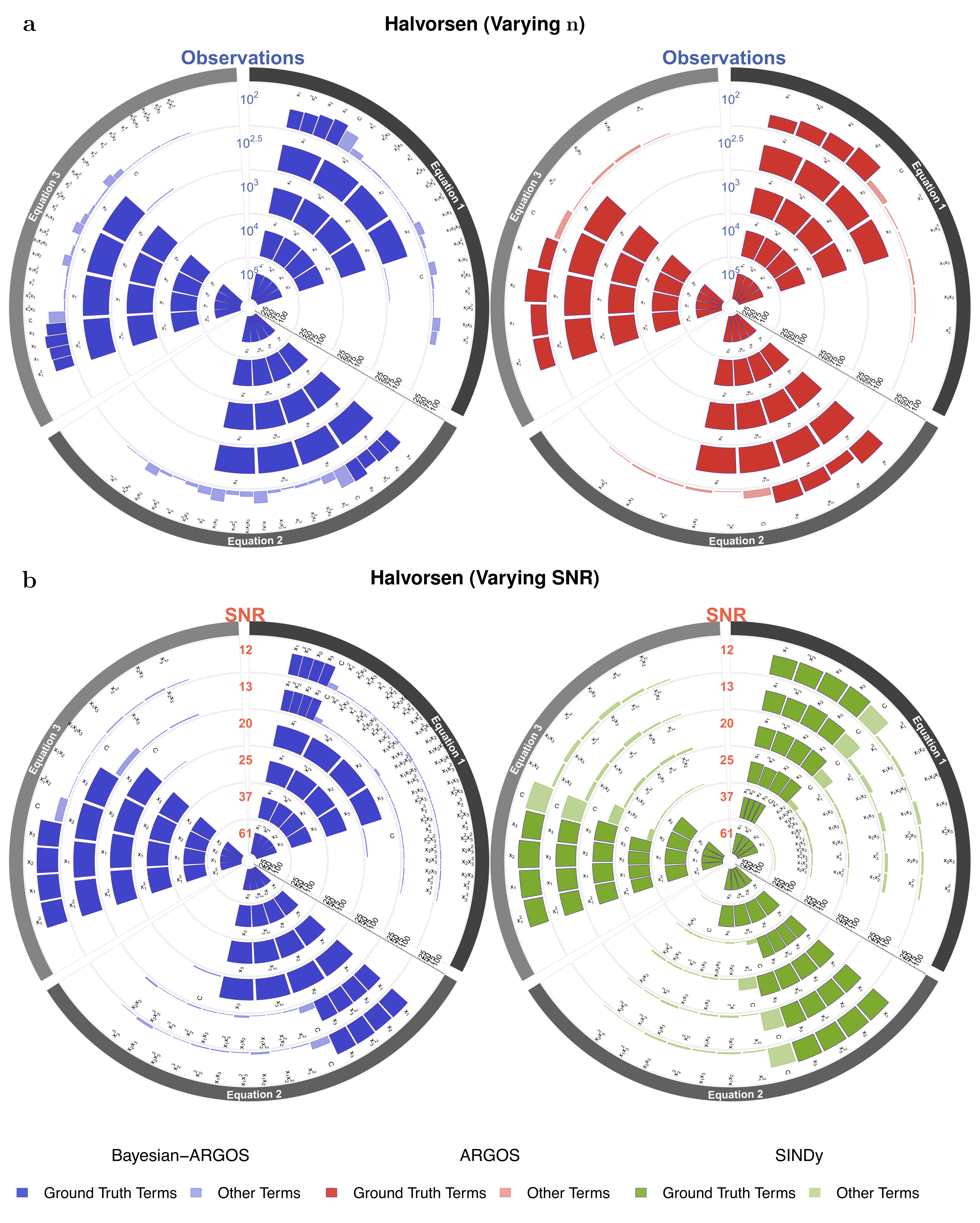}
        \caption{\textbf{Term-selection frequency on the Halvorsen system.} Radial bars show how often each library term is selected across trials for each equation; dark colors denote ground-truth terms and light colors denote other terms. \textbf{a} varying the number of observations; \textbf{b} varying SNR. Bayesian-ARGOS (blue), ARGOS (red), and SINDy (green).}
        \label{fig:si_distribution_halvorsen}
    \end{figure*}

    Figure~\ref{fig:si_distribution_halvorsen} summarizes the frequency with which each candidate term is selected across trials as $n$ and SNR vary. As the number of observations increases (Fig.~\ref{fig:si_distribution_halvorsen}\textbf{a}), Both Bayesian-ARGOS and ARGOS rapidly concentrates selection on the ground-truth support across all three equations. However,Bayesian-ARGOS includes more spurious term activations at $n=10^2$ and a few erroneous constant forcing terms at $n=10^{2.5}$, which should be caused by its balanced shrinkage. As SNR increases (Fig.~\ref{fig:si_distribution_halvorsen}\textbf{b}), Bayesian-ARGOS progressively suppresses spurious activations and stabilizes support recovery. In contrast, SINDy maintains a broader selection distribution, activating numerous non-ground-truth polynomial terms--- at low-to-moderate SNR---even when the dominant ground-truth components are correctly selected.

    \section{Analytical solution of the affine-linear latent dynamics}
    \label{sec:analytic_affine_latent}

    We consider the identified affine-linear latent dynamics (Eq.~\eqref{eq:latent_dynamics_identified}), written in vector form as
    \begin{equation}
        \label{eq:affine_matrix_form}
        \dot z(t)=Az(t)+b,\qquad z(t)\in\mathbb{R}^3,
    \end{equation}
    with

    \begin{equation}
        \label{eq:A_matrix}
        A=
        \begin{pmatrix}
            5.7937848299  & -1.6307242075 & 10.4663248527 \\
            -7.8468539804 & 0             & -8.2580470557 \\
            -8.6063420361 & 0             & -6.6203301732
        \end{pmatrix}, \qquad
        b=
        \begin{pmatrix}
            0.2946496656  \\
            -0.1769868349 \\
            -0.0630564452
        \end{pmatrix}.
    \end{equation}

    A steady state $z^\ast$ satisfies $0=Az^\ast+b$. Since $A$ is invertible for the present model, this equilibrium is unique and given by
    \begin{equation}
        \label{eq:zstar}
        z^\ast=-A^{-1}b.
    \end{equation}
    Introduce centered coordinates
    \begin{equation}
        \label{eq:centered_coords}
        y(t):=z(t)-z^\ast .
    \end{equation}
    Differentiating and substituting \eqref{eq:affine_matrix_form} yields
    \begin{equation}
        \dot y=\dot z = A(y+z^\ast)+b = Ay + (Az^\ast+b)=Ay,
    \end{equation}
    so the dynamics reduce to the homogeneous linear system
    \begin{equation}
        \label{eq:homog}
        \dot y(t)=Ay(t).
    \end{equation}
    For reference, for the identified coefficients one finds $z^\ast\approx (0.036,\,-0.055,\,-0.056)^\top$.

    A standard result for constant-coefficient linear ODEs (equivalently, the fundamental-matrix construction together with existence/uniqueness for linear systems) gives the unique solution of \eqref{eq:homog} as
    \begin{equation}
        \label{eq:matrixexp}
        y(t)=e^{At}y(0),
        \qquad
        e^{At}:=\sum_{k=0}^{\infty}\frac{(At)^k}{k!}.
    \end{equation}
    Reverting the shift \eqref{eq:centered_coords} then gives the explicit solution of the original affine system:
    \begin{equation}
        \label{eq:affine_solution}
        z(t)=z^\ast + e^{At}\bigl(z(0)-z^\ast\bigr).
    \end{equation}

    To express \eqref{eq:matrixexp} in elementary real modes, we use the eigen-structure of $A$. Since $A$ is real, any non-real eigenvalue occurs together with its complex conjugate, and the corresponding eigenvectors are conjugate as well. Assume $A$ has a real eigenvalue $\gamma\in\mathbb{R}$ with a real eigenvector $v_3\in\mathbb{R}^3$, and a complex-conjugate eigenpair $\alpha\pm i\beta$ ($\beta>0$) with eigenvector $v=p+iq,  p,q\in\mathbb{R}^3$,
    satisfying
    \begin{equation}
        A(p+iq)=(\alpha+i\beta)(p+iq).
    \end{equation}
    Then the associated complex mode is $e^{(\alpha+i\beta)t}(p+iq)$. Using Euler's identity
    $e^{i\beta t}=\cos(\beta t)+i\sin(\beta t)$ and taking real and imaginary parts produces two real solutions:
    \begin{equation}
        \begin{aligned}
            \Re\!\left(e^{(\alpha+i\beta)t}(p+iq)\right)
             & =
            e^{\alpha t}\bigl(p\cos(\beta t)-q\sin(\beta t)\bigr),
            \\
            \Im\!\left(e^{(\alpha+i\beta)t}(p+iq)\right)
             & =
            e^{\alpha t}\bigl(p\sin(\beta t)+q\cos(\beta t)\bigr).
        \end{aligned}
    \end{equation}
    Together with the real mode $e^{\gamma t}v_3$, these span the real solution space of \eqref{eq:homog}. Hence the general real solution can be written as
    \begin{equation}
        \label{eq:sincos_general}
        y(t)
        =
        c_1 e^{\gamma t}v_3
        +
        c_2 e^{\alpha t}\bigl(p\cos(\beta t)-q\sin(\beta t)\bigr)
        +
        c_3 e^{\alpha t}\bigl(p\sin(\beta t)+q\cos(\beta t)\bigr),
    \end{equation}
    with constants $c_1,c_2,c_3\in\mathbb{R}$ determined by the initial condition $y(0)=z(0)-z^\ast$.
    Finally,
    \begin{equation}
        \label{eq:final_z_from_y}
        z(t)=z^\ast+y(t).
    \end{equation}
    Equation~\eqref{eq:sincos_general} is equivalent to the more compact representation
    \begin{equation}
        \label{eq:analytical_solution_of_latent_dynamics_2}
        y(t)= c_1 v_{\mathrm{fast}} e^{\gamma t}
        + e^{\alpha t}\Big( c_2 v_{\mathrm{c}}\cos(\beta t) + c_3 v_{\mathrm{s}}\sin(\beta t)\Big),
    \end{equation}
    after a (constant) reparameterization within the two-dimensional real invariant subspace spanned by $\{p,q\}$: one may take $v_{\mathrm{c}}=p$ and $v_{\mathrm{s}}=q$ (up to an invertible linear transformation and corresponding change of coefficients).

    For the identified matrix $A$, spectral analysis yields
    \begin{equation}
        \lambda_{1,2}=\alpha\pm i\beta\approx -0.0124900444 \pm 6.2372886021\,i,
        \qquad
        \lambda_3=\gamma\approx -0.8015652545.
    \end{equation}
    Thus the oscillatory mode has angular frequency $\omega=\beta\approx 6.24~\mathrm{rad\,yr^{-1}}$ and period
    $T=2\pi/\beta\approx 1.01~\mathrm{yr}$, while the decay rates are governed by $\alpha<0$ and $\gamma<0$.
    The oscillatory component is weakly damped with envelope $e^{\alpha t}$ (half-life $\ln 2/|\alpha|$),
    and the fast mode decays on timescale $1/|\gamma|\approx 1.25~\mathrm{yr}$.

    \section{Observed and forecast sea-surface temperature (SST) trajectories at 18 randomly selected held-out sensor locations}

    \begin{sidewaysfigure}
        \centering
        \includegraphics[width=\textheight]{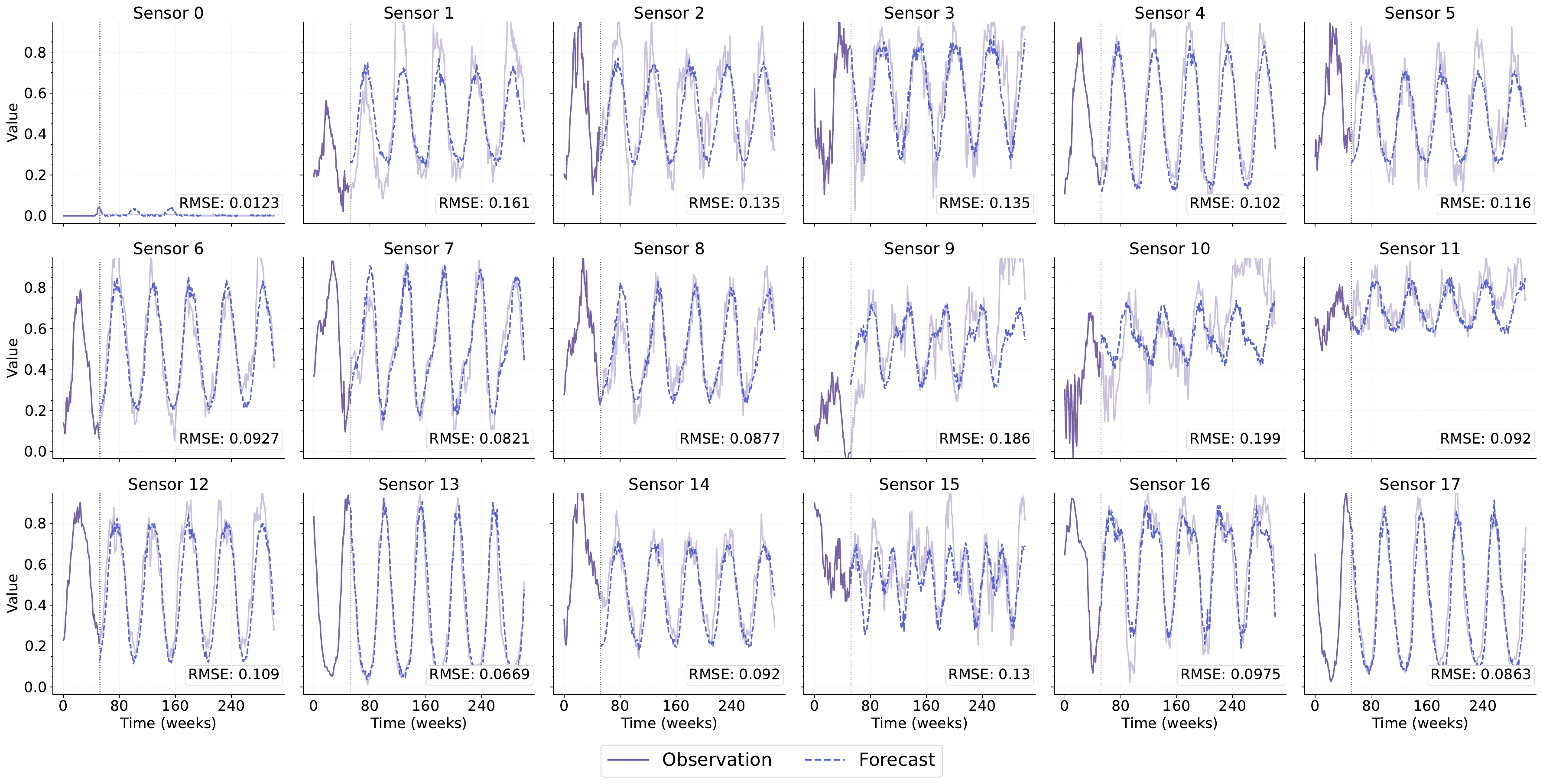}
        \caption{\textbf{Observed and forecast sea-surface temperature (SST) trajectories at 18 randomly selected held-out sensor locations.} Forecasts are obtained by decoding Bayesian-ARGOS simulations of the identified latent dynamics and are shown over the time step from $t=0$ to $t=250$ weeks,  The mean RMSE over this horizon is reported for each location. Observations are shown as solid lines and forecasts as dashed lines.}
        \label{fig:multi_sensor_forecast}
    \end{sidewaysfigure}

    \clearpage

    \section{Algorithms}
    \label{sec:algorithms}

    \begin{algorithm*}[!htp]
        \caption{Adaptive Savitzky-Golay Filter}
        \label{al:SG}
        \begin{algorithmic}[1]
            \Require $\mathbf{\Tilde{X}}\in \reals^{n\times m}$, $\ dt$.
            \Ensure Savitzky-Golay optimally smoothed $\mathbf{X}$ and $\mathbf{\dot{X}}$.
            \\

            Determine the admissible (odd) window-length range:
            $l_{\min}=13$ and
            $l_{\max}=\max\!\left(13,\ \min\!\left(n-(n-1)\bmod 2,\ 101\right)\right)$;\\

            Construct the candidate set of odd window lengths
            $L=\{l_{\min},\,l_{\min}+2,\,\dots,\,l_{\max}\}$;

            \color{blue}
            \LeftComment{$v$ denotes the derivative order in $SG(\cdot)$}
            \normalcolor
            \For{$j=1,\dots,m$}

            Let $\mathbf{\Tilde{x}}_j$ be the $j$th column of $\mathbf{\Tilde{X}}$;

            $\begin{aligned}
                    l^{\ast} & =\arg\min_{l\in L}\left\|SG(\mathbf{\Tilde{x}}_j,\,o=4,\,l=l,\,v=0,\,dt)-\mathbf{\Tilde{x}}_j\right\|_2^2; \\
                \end{aligned}$

            $\begin{aligned}
                    \mathbf{x}_j=SG(\mathbf{\Tilde{x}}_j,\,o=4,\,l=l^{\ast},\,v=0,\,dt);
                \end{aligned}$

            $\begin{aligned}
                    \mathbf{\dot{x}}_j=SG(\mathbf{\Tilde{x}}_j,\,o=4,\,l=l^{\ast},\,v=1,\,dt);
                \end{aligned}$
            \EndFor\\

            Assemble $\mathbf{X}=[\mathbf{x}_1,\dots,\mathbf{x}_m]\in\reals^{n\times m}$ and
            $\mathbf{\dot{X}}=[\mathbf{\dot{x}}_1,\dots,\mathbf{\dot{x}}_m]\in\reals^{n\times m}$.
        \end{algorithmic}
    \end{algorithm*}

    \begin{algorithm}[!htp]
        \caption{Bayesian Regression-based Automatic Regression for Governing Equations (Bayesian-ARGOS)}
        \label{al:Bayesian_ARGOS}

        \begin{algorithmic}[1]
            \Require $\mathbf{X}\in \mathbb{R}^{n\times m}$, $\mathbf{\dot{X}}\in \mathbb{R}^{n\times m}$, polynomial degree $d_1$ and number of domain-specific functions $d_2$, credible level $(1-\alpha)$ with $\alpha=0.10$, adaptive-LASSO tuning $\gamma>0$.
            \Ensure Retained terms for each component equation $\dot{\mathbf{x}}_j$, $j=1,\dots,m$.
            \For{$j=1,\dots,m$}
            Set response $\mathbf{y}=\dot{\mathbf{x}}_j\in\mathbb{R}^{n}$.

            \color{blue}
            \LeftComment{STEP ONE: Initial design matrix}
            \normalcolor\\
            $p^{(0)}=\binom{m+d_1}{d_1} + d_2$;\\
            Construct $\mathbf{\Theta}^{(0)}(\mathbf{X})\in\mathbb{R}^{n\times p^{(0)}}$ with monomials up to degree $d_1$ of the columns of $\mathbf{X}$ and include $d_2$ domain-specific functions;

            \color{blue}
            \LeftComment{STEP TWO: First-pass screening (two-stage procedure)}
            \normalcolor

            \color{blue}
            % \Statex \hspace{\algorithmicindent} \(\triangleright\) 
            \LeftComment{Ridge-weighted Adaptive LASSO}
            \normalcolor\\

            Compute Ridge pilot estimates $\hat{\boldsymbol{\beta}}^{\text{ridge}}$ by regressing $\mathbf{y}$ on $\mathbf{\Theta}^{(0)}(\mathbf{X})$; set adaptive weights $w_k=\left|\hat{\beta}^{\text{ridge}}_{k}\right|^{-\gamma}$;\label{algo_step:first_stage_regression}\\
            Choose $\lambda^*$ from two-phase grid search;\\
            $\begin{aligned}
                    \hat{\boldsymbol{\beta}}^{(1)}
                    =\arg\min_{\boldsymbol{\beta}} \left\|\mathbf{y}-\mathbf{\Theta}^{(0)}(\mathbf{X})\boldsymbol{\beta}\right\|_2^2
                    +\lambda^*\sum_{k=1}^{p^{(0)}} w_k |\beta_k|;
                \end{aligned}$

            \color{blue}
            % \Statex \hspace{\algorithmicindent} \(\triangleright\) 
            \LeftComment{Ordinary least squares (OLS) refit and BIC selection over thresholds}
            \normalcolor\\
            $\begin{aligned}\eta = [10^{-8},10^{-7},\dotsc, 10^{1}]\end{aligned}$;

            \For{$i=1,\dots,\mathrm{card}(\eta)$}

            $\begin{aligned} \hat{\boldsymbol{\beta}}^{\text{OLS}(1)}[i]
                    =\arg\min_{\boldsymbol{\beta}_{\mathcal{K}_i}}
                    \left\|\mathbf{y}-\mathbf{\Theta}^{(0)}_{\mathcal{K}_i}(\mathbf{X})\boldsymbol{\beta}_{\mathcal{K}_i}\right\|_2^2;\ \text{where}\ \mathcal{K}_i=\{k:\left|\hat{\beta}^{(1)}_{k}\right|\ge \eta_i\};\end{aligned}$

            $\text{BIC}_i=\text{BIC}(\hat{\boldsymbol{\beta}}^{\text{OLS}(1)}[i])$;
            \EndFor\\
            Obtain $\mathcal{K}^{(1)}=\mathcal{K}_{i^*}$, where $i^*=\arg\min_i \text{BIC}_i$;  \label{algo_step:second_stage_regression}
            \color{blue}
            \LeftComment{Strategic refinement of candidate library}
            \normalcolor\\
            Construct refined library $\mathbf{\Theta}^{(1)}(\mathbf{X})$ with basis functions up to degree $d^{(1)}$, where $d^{(1)} = \max(\text{degree}(k) \mid k \in \mathcal{K}^{(1)})$;
            \color{blue}
            \LeftComment{STEP THREE: Second-pass screening}
            \normalcolor\\
            Repeat Statements~\ref{algo_step:first_stage_regression}~--~\ref{algo_step:second_stage_regression} with OLS pilot estimates $\hat{\boldsymbol{\beta}}^{\text{OLS}}$ on $\mathbf{\Theta}^{(1)}(\mathbf{X})$, set $w_k=\left|\hat{\beta}^{\text{OLS}}_{k}\right|^{-\gamma}$, and obtain $\mathcal{K}^{(2)}$;\\
            Obtain trimmed library $\mathbf{\Theta}^{(2)}(\mathbf{X})=\mathbf{\Theta}^{(1)}_{\mathcal{K}^{(2)}}(\mathbf{X})$;

            \color{blue}
            \LeftComment{STEP FOUR: Bayesian regression and uncertainty quantification}
            \normalcolor\\
            Model $\mathbf{y} \sim \mathcal{N}(\mathbf{\Theta}^{(2)}(\mathbf{X})\boldsymbol{\beta},\sigma^2\mathbf{I}_n)$ with priors $\beta_{k}\sim \mathcal{N}\!\left(0,(2.5s_y/s_{x_k})^2\right)$,\quad $\sigma\sim \text{Exp}(1/s_y)$, where $s_y=\mathrm{sd}(\mathbf{y})$ and $s_{x_k}=\mathrm{sd}(\mathbf{\Theta}^{(2)}_k)$;\\
            Sample posterior $p(\boldsymbol{\beta}, \sigma \mid \mathbf{\Theta}^{(2)}(\mathbf{X}), \mathbf{y})$ via HMC (4 chains, 2000 iterations each, 1000 warmup);\\
            Compute posterior mean $\mu_k$ and $(1-\alpha)$ credible interval $[CI^{\text{lo}}_k,CI^{\text{up}}_k]$ conditioned on $\mathbf{\Theta}^{(2)}(\mathbf{X})$ for each $\beta_k$;\\
            Retain term $k$ iff $0 \notin [CI^{\text{lo}}_k,CI^{\text{up}}_k]$.
            \EndFor
        \end{algorithmic}
    \end{algorithm}

    \clearpage
    \begingroup
    \renewcommand{\refname}{Supplementary References}
    \putbib[references1,references2]
    \endgroup
\end{bibunit}

\end{document}